\documentclass[10pt,twocolumn,letterpaper]{article}

\usepackage[pagenumbers]{cvpr} 
\usepackage{xcolor}
\usepackage{kotex}
\usepackage{multirow}
\usepackage{graphicx}
\usepackage{bbm}

\usepackage{times}
\usepackage{epsfig}
\usepackage{graphicx}
\usepackage{amsmath}
\usepackage{amssymb}
\usepackage{nicefrac}
\usepackage{pifont}
\usepackage{dsfont}
\usepackage{tabularx}
\usepackage{tabulary}
\usepackage{booktabs}
\usepackage{multirow}
\usepackage{subcaption}
\usepackage{colortbl}

\newcommand{\figref}[1]{Figure~\ref{fig:#1}}
\newcommand{\tabref}[1]{Table~\ref{tab:#1}} 
\newcommand{\eqnref}[1]{Eqn.~\eqref{eq:#1}}
\newcommand{\secref}[1]{Section~\ref{sec:#1}}

\definecolor{tabzeroth}{rgb}{1, 0.5, 0.5}
\definecolor{tabfirst}{rgb}{1, 0.7, 0.7} 
\definecolor{tabsecond}{rgb}{1, 0.85, 0.7}
\definecolor{tabthird}{rgb}{1, 1, 0.7} 

\definecolor{cvprblue}{rgb}{0.21,0.49,0.74}
\usepackage[pagebackref,breaklinks,colorlinks,citecolor=cvprblue]{hyperref}

\def\paperID{14196} 
\def\confName{CVPR}
\def\confYear{2024}

\title{Depth-Regularized Optimization for 3D Gaussian Splatting in Few-Shot Images}

\author{Jaeyoung Chung$^{1}$ \qquad\qquad\qquad Jeongtaek Oh$^{2}$ \qquad\qquad\qquad Kyoung Mu Lee$^{1,2}$\\
$^{1}$Department of ECE, ASRI, Seoul National University, Seoul, Korea\\
$^{2}$IPAI, ASRI, Seoul National University, Seoul, Korea\\
{\tt\small \{robot0321, ohjtgood, kyoungmu\}@snu.ac.kr}
}

\begin{document}
\maketitle
\begin{abstract}
In this paper, we present a method to optimize Gaussian splatting with a limited number of images while avoiding overfitting.
Representing a 3D scene by combining numerous Gaussian splats has yielded outstanding visual quality.
However, it tends to overfit the training views when only a small number of images are available.
To address this issue, we introduce a dense depth map as a geometry guide to mitigate overfitting.
We obtained the depth map using a pre-trained monocular depth estimation model and aligning the scale and offset using sparse COLMAP feature points.
The adjusted depth aids in the color-based optimization of 3D Gaussian splatting, mitigating floating artifacts, and ensuring adherence to geometric constraints.
We verify the proposed method on the NeRF-LLFF dataset with varying numbers of few images.
Our approach demonstrates robust geometry compared to the original method that relies solely on images.
\end{abstract}    
\section{Introduction} \label{sec:intro}

\begin{figure}[t!]
    \centering
    \includegraphics[width=0.48\textwidth, height=0.48\textwidth]{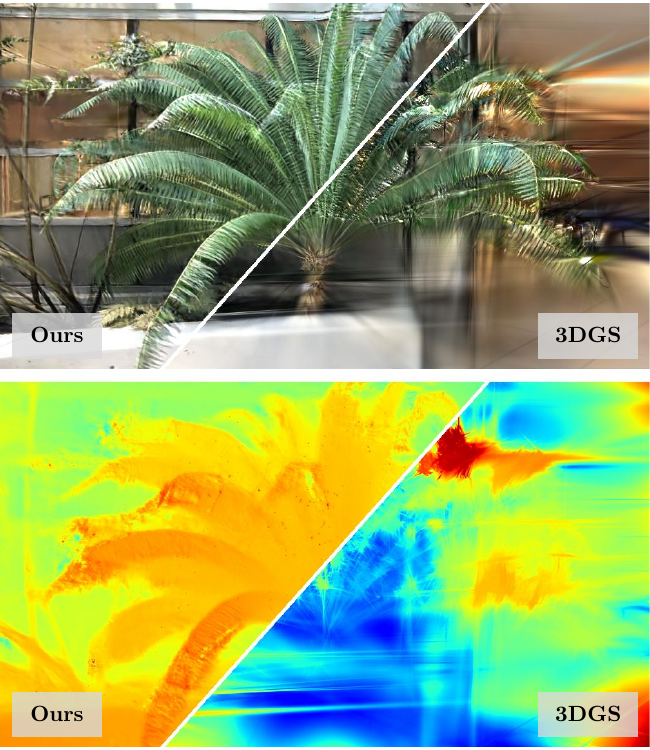}
    \vspace{-0.6cm}
    \caption{\textbf{The efficacy of depth regularization in a few-shot setting} We optimize Gaussian splats with a limited number of images, avoiding overfitting through the geometry guidance estimated from the images. Please note that we utilized only \textit{two} images to create this 3D scene.
    }
    \label{fig:teaser}
    \vspace{-0.6cm}
\end{figure}

Reconstruction of three-dimensional space from images has long been a challenge in the computer vision field. 
Recent advancements show the feasibility of photorealistic novel view synthesis~\cite{barron2021mip, mildenhall2021nerf}, igniting research into reconstructing a complete 3D space from images. 
Driven by progress in computer graphics techniques and industry demand, particularly in sectors such as virtual reality~\cite{deng2022fov} and mobile~\cite{chen2023mobilenerf}, research on achieving high-quality and high-speed real-time rendering has been ongoing.
Among the recent notable developments, 3D Gaussian Splatting (3DGS)~\cite{kerbl20233d} stands out through its combination of high quality, rapid reconstruction speed, and support for real-time rendering. 
3DGS employs Gaussian attenuated spherical harmonic splats~\cite{reed2012methods, crawfis1993texture} with opacity as primitives to represent every part of a scene.
It guides the splats to construct a consistent geometry by imposing a constraint on the splats to satisfy multiple images at the same time.
%

The approach of aggregating small splats for a scene provides the capability to express intricate details, yet it is prone to overfitting due to its local nature. 
3DGS~\cite{kerbl3Dgaussians} optimizes independent splats according to multi-view color supervision without global structure.
Therefore, in the absence of a sufficient quantity of images that can offer a global geometric cue, there exists no precaution against overfitting.
This issue becomes more pronounced as the number of images used for optimizing a 3D scene is small. 
The limited geometric information from a few number of images leads to an incorrect convergence toward a local optimum, resulting in optimization failure or floating artifacts as shown in \figref{teaser}.
Nevertheless, the capability to reconstruct a 3D scene with a restricted number of images is crucial for practical applications, prompting us to tackle the few-shot optimization problem.

One intuitive solution is to supplement an additional geometric cue such as depth.
In numerous 3D reconstruction contexts~\cite{bian2022nopenerf}, depth proves immensely valuable for reconstructing 3D scenes by providing direct geometric information.
To obtain such robust geometric cues, depth sensors aligned with RGB cameras are employed.
Although these devices offer dense depth maps with minimal error, the necessity for such equipment also presents obstacles to practical applications.

Hence, we attain a dense depth map by adjusting the output of the depth estimation network with a sparse depth map from the renowned Structure-from-Motion (SfM), which computes the camera parameters and 3D feature points simultaneously.
3DGS also uses SfM, particularly COLMAP~\cite{schonberger2016structure}, to acquire such information.
However, the SfM also encounters a notable scarcity in the available 3D feature points when the number of images is few.
The sparse nature of the point cloud also makes it impractical to regularize all Gaussian splats.
Hence, a method for inferring dense depth maps is essential.
One of the methods to extract dense depth from images is by utilizing monocular depth estimation models.
While these models are able to infer dense depth maps from individual images based on priors obtained from the data, they produce only relative depth due to scale ambiguity.
Since the scale ambiguity leads to critical geometry conflicts in multi-view images, we need to adjust scales to prevent conflicts between independently inferred depths. We show that this can be done by fitting a sparse depth, which is a free output from COLMAP~\cite{schonberger2016structure} to an estimated dense depth map.

In this paper, we propose a method to represent 3D scenes using a small number of RGB images leveraging prior information from a pre-trained monocular depth estimation model~\cite{bhat2023zoedepth} and a smoothness constraint. 
We adapt the scale and offset of the estimated depth to the sparse COLMAP points, solving the scale ambiguity. 
We use the adjusted depth as a geometry guide to assist color-based optimization, reducing floating artifacts and satisfying geometry conditions. 
We observe that even the revised depth helps guide the scene to geometrically optimal solution despite its roughness.
We prevent the overfitting problem by incorporating an early stop strategy, where the optimization process stops when the depth-guide loss starts to rise.
Moreover, to achieve more stability, we apply a smoothness constraint, ensuring that neighbor 3D points have similar depths.
We adopt 3DGS as our baseline and compare the performance of our method in the NeRF-LLFF~\cite{mildenhall2019llff} dataset.
We confirm that our strategy leads to plausible results not only in terms of RGB novel-view synthesis but also 3D geometry reconstruction. 
Through further experiments, we demonstrate the influence of geometry cues such as depth and initial points on Gaussian splatting.
They significantly influence the stable optimization of Gaussian splatting.

In summary, our contributions are as follows: 
\begin{description}
\item[$\bullet$] We propose depth-guided Gaussian Splatting optimization strategy which enables optimizing the scene with a few images, mitigating over-fitting issue. We demonstrate that even an estimated depth adjusted with a sparse point cloud, which is an outcome of the SfM pipeline, can play a vital role in geometric regularization.

\item[$\bullet$] We present a novel early stop strategy: \textit{halting} the training process when depth-guided loss suffers to drop. We illustrate the influence of each strategy through thorough ablation studies. 

\item[$\bullet$] We show that the adoption of a smoothness term for the depth map directs the model to finding the correct geometry. Comprehensive experiments reveal enhanced performance attributed to the inclusion of a smoothness term.
\end{description}
\section{Related Work} \label{sec:relatedwork}
\paragraph{Novel view synthesis}
Structure from motion 
(SfM)~\cite{ullman1979interpretation} and Multi-view stereo (MVS)~\cite{tomasi1992shape} are techniques for reconstructing 3D structures using multiple images, which have been studied for a long time in the computer vision field.
Among the continuous developments, COLMAP~\cite{schonberger2016structure} is a widely used representative tool.
COLMAP performs camera pose calibration and finds sparse 3D keypoints using the epipolar constraint~\cite{hartley2003multiple} of multi-view images.
For more dense and realistic reconstruction, deep learning based 3D reconstruction techniques have been mainly studied.~\cite{han2019image, xie2022neural, mildenhall2021nerf}
Among them, Neural radiance fields (NeRF)~\cite{mildenhall2021nerf} is a representative method that uses a neural network as a representation method.
NeRF creates realistic 3D scenes using an MLP network as a 3D space expression and volume rendering, producing many follow-up papers on 3D reconstruction research.~\cite{tewari2022advances, gao2022nerf, barron2021mip, yariv2021volume, wang2021neus, barron2022mip}
In particular, to overcome slow speed of NeRF, many efforts continues to achieve real-time rendering by utilizing explicit expression such as sparse voxels\cite{liu2020neural, yu2021plenoctrees, sun2022direct, fridovich2022plenoxels}, featured point clouds\cite{xu2022point}, tensor~\cite{chen2022tensorf}, polygon~\cite{chen2023mobilenerf}.
These representations have local elements that operate independently, so they show fast rendering and optimization speed.
Based on this idea, various representations such as Multi-Level Hierarchies~\cite{muller2021real, muller2022instant}, infinitesimal networks~\cite{garbin2021fastnerf, reiser2021kilonerf}, triplane~\cite{chan2022efficient} have been attempted.
Among them, 3D Gaussian splatting~\cite{kerbl20233d} presented a fast and efficient method through alpha-blending rasterization instead of time-consuming volume rendering. 
It optimizes a 3D scene using multi-million Gaussian attenuated spherical harmonics with opacity as a primitive, showing easy and fast 3D reconstruction with high quality.
 
\begin{figure*}[t!]
    \centering
    \includegraphics[width=0.98\textwidth]{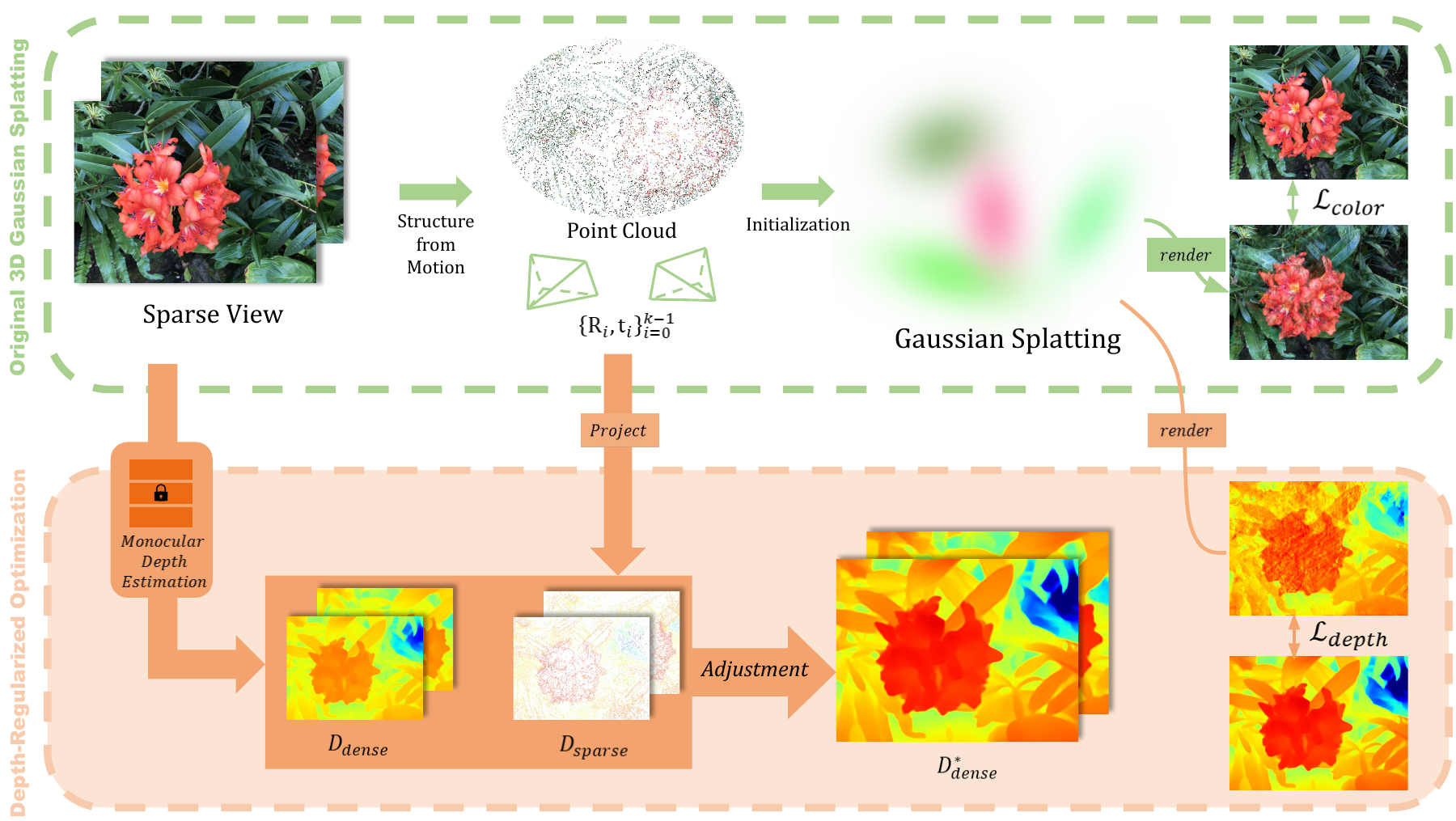}
    \vspace{-0.2cm}
    \caption{\textbf{Overview.}
    We optimize the 3D Gaussian splatting~\cite{kerbl20233d} using dense depth maps adjusted with the point clouds obtained from COLMAP~\cite{schonberger2016structure}. 
    By incorporating depth maps to regulate the geometry of the 3D scene, our model successfully reconstructs scenes using a limited number of images.
    }
    \label{fig:method}
    \vspace{-0.4cm}
\end{figure*}

\paragraph{Few-shot 3D reconstruction}
Since an image contains only partial information about the 3D scene, 3D reconstruction requires a large number of multi-view images.
COLMAP uploads feature points matched between multiple images onto 3D space, so the more images are used, the more reliable 3D points and camera poses can be obtained.\cite{schonberger2016structure, furukawa2015multi}
NeRF also optimizes the color and geometry of a 3D scene based on the pixel colors of a large number of images to obtain high-quality scenes.~\cite{wang2021nerf, zhang2020nerf++}
However, the requirements for many images hindered practical application, sparking research on 3D reconstruction using only a few number of images.
Many few-shot 3D reconstruction studies utilize depth to provide valuable geometric cue for creating 3D scenes.
Depth helps reduce the effort of inferring geometry through color consensus in multiple images by introducing a surface smoothness constraint~\cite{kim2022infonerf, niemeyer2022regnerf}, supervising the sparse depth obtained from COLMAP~\cite{deng2022depth, wei2021nerfingmvs}, using the dense depth obtained from additional sensors~\cite{azinovic2022neural, cai2022neural, dey2022mip}, or exploiting estimated dense depth from pretrained network.~\cite{prinzler2023diner, roessle2022dense, neff2021donerf}
These studies regularize geometry based on the globality of the neural network, so it is difficult to apply them to representations with large locality such as sparse voxel~\cite{fridovich2022plenoxels} or feature point~\cite{xu2022point}.
Instead, they attempted to establish connectivity between local elements in a 3D space through the total variation (TV) loss~\cite{zhou2017unsupervised, fridovich2022plenoxels, yang2023nerfvs}, but it requires exhaustive hyperparameter tuning of total variation which varies on the scene and location.
3D Gaussian splatting~\cite{kerbl20233d} generates floating artifacts with a few number of images, due to its strong locality.
The sparse COLMAP feature points that can be obtained from the Gaussian splat subprocess are a free depth guide that can be obtained without additional information~\cite{roessle2022dense}, but the number of sparse points obtained from a small number of images is so small that it cannot guide all Gaussian splats with strong locality.
We use a coarse geometry guide for optimization through a pretrained depth estimation model~\cite{bhat2023zoedepth, zhao2023unleashing, mertan2022single}. 
Even if they do not have an exact fine-detailed depth, they provide a rough guide to the location of splats, which greatly contributes to optimization stability in few-shot situations and helps eliminate floating artifacts that occur in random locations.

\section{Method} \label{sec:method}
Our method facilitates the optimization from a small set of images $\{I_i\}_{i=0}^{k-1}, I_i \in [0,1]^{H\times W \times 3}$. 
As a preprocessing, we run SfM (such as COLMAP\cite{schonberger2016structure}) pipeline and get the camera pose $\mathrm{R_i}\in\mathbb{R}^{3\times3}, \mathrm{t_i}\in\mathbb{R}^3$, intrinsic parameters $K_i\in\mathbb{R}^{3\times3}$, and a point cloud $P \in \mathbb{R}^{n \times 3}$. With those informations, we can easily obtain a sparse depth map for each image, by projecting all visible point to pixel space:
\begin{gather}
    p = {P_{\text{homog}}}[\mathrm{R_i}|\mathrm{t_i}],  \\
    \text{and} \quad D_{sparse,i} = p_z  \in [0,\infty]^{H\times W}.
\end{gather}    
Our approach builds upon 3DGS~\cite{kerbl20233d}.
They optimize the Gaussian splats based on the rendered image with a color loss $\mathcal{L}_{color}$ and D-SSIM loss $\mathcal{L}_{D-SSIM}$. 
Prior to the 3DGS optimization, we estimate a depth map for each image using a depth estimation network and fit the sparse depth map\secref{3_EstDepthOpt}).
We render a depth from the set of Gaussian splattings leveraging the color rasterization process and add a depth constraint using the dense depth prior (\secref{3_RenderDepthFit}).
We add an additional constraint for smoothness between depths of adjacent pixels (\secref{3_DepthSmooth}) and refine optimization options for few-shot settings (\secref{3_OptSet}).

\subsection{Preparing Dense Depth Prior} \label{sec:3_EstDepthOpt}
With the goal of guiding the splats into plausible geometry, we require to provide global geometry information due to the locality of Gaussian splats.
The density depth is one of the promising geometry prior, but there is a challenge in constructing it.
The density of SfM points depends on the number of images, so the number of valid points are too small to directly estimate dense depth in a few-shot setting.
(For example, SfM reconstruction from 19 images creates a sparse depth map with 0.04\% valid pixels on average.~\cite{roessle2022dense})
Even the latest depth completion models fail to complete dense depth due to the significant information gap.

When designing the depth prior, it is important to note that even rough depth significantly aids in guiding the splats and eliminating artifacts resulting from splats trapped in incorrect geometry.
Hence, we employ a state-of-the-art monocular depth estimation model and scale matching to provide a coarse dense depth guide for optimization. 
From a train image $I$, the monocular depth estimation model $F_{\theta}$ outputs dense depth $D_{\text{dense}}$,
\begin{align}
    D_{\text{dense}} = s \cdot F_{\theta}(I)+t.
\end{align}
To resolve the scale ambiguity in the estimated dense depth $D_{\text{dense}}$, we adjust the scale $s$ and offset $t$ of estimated depth to sparse SfM depth $D_{\text{sparse}}$:
\begin{align}
    s^{\ast},t^{\ast} = \arg\min\limits_{s,t}\sum_{p\in D_{\text{sparse}}} {\begin{Vmatrix}w(p){\cdot}D_{\text{sparse}}(p) - D_{\text{dense}}(p;s,t)\end{Vmatrix}^2},
    \label{eq:fitting_estdepth}
\end{align}
where $w\in[0,1]$ is a normalized weight presenting the reliability of each feature points calculated as the reciprocal of the reprojection error from SfM.
Finally, we use the adjusted dense depth $D^{\ast}_{\text{dense}}= s^{\ast}\cdot F_{\theta}(I)+t^{\ast}$ to regularize the optimization loss of Gaussian splatting.

\subsection{Depth Rendering through Rasterization} \label{sec:3_RenderDepthFit}
3D Gaussian splatting utilizes a rasterization pipeline~\cite{akenine2019real} to render the disconnected and unstructured splats leveraged on the parallel architecture of GPU. 
Based on differentiable point-based rendering techniques~\cite{wiles2020synsin, yifan2019differentiable, kopanas2021point}, they render an image by rasterizing the splats through $\alpha$-blending. 
Point-based approaches exploit a similar equation to NeRF-style volume rendering, rasterizing a pixel color with ordered points that cover that pixel,
\begin{gather}
    \label{eq:color_render}
    C = \sum_{i\in\textit{N}} {c_i \alpha_i T_i} \\
    \text{where}\quad {T_i = \prod_{j=1}^{i-1} (1-\alpha_j)}, \nonumber
\end{gather}    
$C$ is the pixel color, $c$ is the color of splats, and $\alpha$ here is learned opacity multiplied by the covariance of 2D Gaussian.
This formulation prioritizes the color $c$ of opaque splat positioned closer to the camera, significantly impacting the final outcome $C$.
Inspired by the depth implementation in NeRF, we leverage the rasterization pipeline to render the depth map of Gaussian splats,
\begin{align}
    D = \sum_{i\in\textit{N}} {d_i \alpha_i T_i}, \label{eq:depth_render}
\end{align}
where $D$ is the rendered depth and $d_i=(R_i p_i+T_i)_{z}$ is the depth of each splat from the camera. 
\eqnref{depth_render} enables the direct utilization of $\alpha_i$ and $T_i$ calculated in \eqnref{color_render}, facilitating rapid depth rendering with minimal computational load.
Finally, we guide the rendered depth to the estimated dense depth using L1 distance,
\begin{align}
    \mathcal{L}_{depth}=\begin{Vmatrix}D - D^{\ast}_{\text{dense}} \end{Vmatrix}_1 
    \label{eq:depthloss}
\end{align}

\begin{table*}[t]
\centering
\resizebox{\textwidth}{!}{%
\begin{tabular}{|ccc|cccc|cccc|cccc|}
\hline
\multicolumn{3}{|c|}{\multirow{2}{*}{}} &
  \multicolumn{4}{c|}{{PSNR$\uparrow$}} &
  \multicolumn{4}{c|}{{SSIM$\uparrow$}} &
  \multicolumn{4}{c|}{{LPIPS$\downarrow$}} \\
\multicolumn{3}{|c|}{} &
  2-view &
  3-view &
  4-view &
  5-view &
  2-view &
  3-view &
  4-view &
  5-view &
  2-view &
  3-view &
  4-view &
  5-view \\ \hline
\multicolumn{1}{|c|}{\multirow{24}{*}{\rotatebox[origin=c]{90}{NeRF-LLFF\cite{mildenhall2021nerf}}}} &
\multicolumn{1}{|c|}{\multirow{3}{*}{Fern}} &
  3DGS &
  13.03 &
  14.29 &
  16.73 &
  18.59 &
  0.336 &
  0.408 &
  0.517 &
  0.603 &
  0.476 &
  0.389 &
  0.296 &
  0.217 \\
\multicolumn{1}{|c|}{} &
\multicolumn{1}{|c|}{} &
  Ours &
  \textbf{17.59} &
  \textbf{19.13} &
  \textbf{19.91} &
  \textbf{20.55} &
  \textbf{0.516} &
  \textbf{0.588} &
  \textbf{0.616} &
  \textbf{0.642} &
  \textbf{0.286} &
  \textbf{0.232} &
  \textbf{0.203} &
  \textbf{0.167} \\
\multicolumn{1}{|c|}{} &
\multicolumn{1}{|c|}{} &
  Oracle &
  18.18 &
  20.30 &
  20.78 &
  21.81 &
  0.524 &
  0.636 &
  0.654 &
  0.701 &
  0.278 &
  0.201 &
  0.185 &
  0.157 \\  \cline{2-15}
\multicolumn{1}{|c|}{} &
\multicolumn{1}{|c|}{\multirow{3}{*}{Flower}} &
  3DGS &
  14.90 &
  17.75 &
  \textbf{19.71} &
  \textbf{21.39} &
  0.351 &
  \textbf{0.508} &
  \textbf{0.605} &
  \textbf{0.671} &
  \textbf{0.406} &
  \textbf{0.257} &
  \textbf{0.190} &
  \textbf{0.146} \\
\multicolumn{1}{|c|}{} &
\multicolumn{1}{|c|}{} &
  Ours &
  \textbf{15.92} &
  \textbf{17.80} &
  19.15 &
  20.45 &
  \textbf{0.395} &
  0.445 &
  0.538 &
  0.576 &
  0.414 &
  0.376 &
  0.323 &
  0.293 \\
\multicolumn{1}{|c|}{} &
\multicolumn{1}{|c|}{} &
  Oracle &
  19.71 &
  22.16 &
  23.26 &
  24.65 &
  0.570 &
  0.673 &
  0.714 &
  0.760 &
  0.250 &
  0.163 &
  0.128 &
  0.097 \\  \cline{2-15}
\multicolumn{1}{|c|}{} &
\multicolumn{1}{|c|}{\multirow{3}{*}{Fortress}} &
  3DGS &
  13.87 &
  15.98 &
  19.26 &
  19.98 &
  0.363 &
  0.492 &
  0.609 &
  0.631 &
  0.389 &
  0.283 &
  0.201 &
  0.191 \\
\multicolumn{1}{|c|}{} &
\multicolumn{1}{|c|}{} &
  Ours &
  \textbf{19.80} &
  \textbf{21.85} &
  \textbf{23.07} &
  \textbf{23.72} &
  \textbf{0.567} &
  \textbf{0.655} &
  \textbf{0.724} &
  \textbf{0.740} &
  \textbf{0.232} &
  \textbf{0.191} &
  \textbf{0.162} &
  \textbf{0.144} \\
\multicolumn{1}{|c|}{} &
\multicolumn{1}{|c|}{} &
  Oracle &
  23.07 &
  24.51 &
  26.39 &
  26.73 &
  0.654 &
  0.728 &
  0.787 &
  0.797 &
  0.159 &
  0.130 &
  0.100 &
  0.093 \\ \cline{2-15}
\multicolumn{1}{|c|}{} &
\multicolumn{1}{|c|}{\multirow{3}{*}{Horns}} &
  3DGS &
  11.43 &
  12.48 &
  13.76 &
  14.75 &
  0.264 &
  0.339 &
  0.433 &
  0.498 &
  0.531 &
  0.464 &
  0.395 &
  0.350 \\
\multicolumn{1}{|c|}{} &
\multicolumn{1}{|c|}{} &
  Ours &
  \textbf{15.91} &
  \textbf{16.22} &
  \textbf{18.09} &
  \textbf{18.39} &
  \textbf{0.420} &
  \textbf{0.466} &
  \textbf{0.527} &
  \textbf{0.565} &
  \textbf{0.362} &
  \textbf{0.349} &
  \textbf{0.306} &
  \textbf{0.296} \\
\multicolumn{1}{|c|}{} &
\multicolumn{1}{|c|}{} &
  Oracle &
  18.56 &
  20.08 &
  20.88 &
  22.52 &
  0.568 &
  0.644 &
  0.668 &
  0.725 &
  0.259 &
  0.212 &
  0.199 &
  0.168 \\ \cline{2-15}
\multicolumn{1}{|c|}{} &
\multicolumn{1}{|c|}{\multirow{3}{*}{Leaves}} &
  3DGS &
  12.33 &
  12.36 &
  12.49 &
  12.26 &
  \textbf{0.260} &
  \textbf{0.275} &
  \textbf{0.298} &
  \textbf{0.297} &
  \textbf{0.412} &
  \textbf{0.397} &
  \textbf{0.397} &
  \textbf{0.401} \\
\multicolumn{1}{|c|}{} &
\multicolumn{1}{|c|}{} &
  Ours &
  \textbf{13.04} &
  \textbf{13.63} &
  \textbf{13.97} &
  \textbf{14.13} &
  0.235 &
  0.270 &
  0.283 &
  \textbf{0.297} &
  0.460 &
  0.445 &
  0.440 &
  0.438 \\
\multicolumn{1}{|c|}{} &
\multicolumn{1}{|c|}{} &
  Oracle &
  13.52 &
  14.23 &
  14.78 &
  14.85 &
  0.287 &
  0.353 &
  0.377 &
  0.397 &
  0.380 &
  0.348 &
  0.341 &
  0.356 \\ \cline{2-15}
\multicolumn{1}{|c|}{} &
\multicolumn{1}{|c|}{\multirow{3}{*}{Orchids}} &
  3DGS &
  11.78 &
  13.94 &
  \textbf{15.41} &
  16.08 &
  0.182 &
  \textbf{0.320} &
  \textbf{0.416} &
  \textbf{0.460} &
  \textbf{0.426} &
  \textbf{0.310} &
  \textbf{0.245} &
  \textbf{0.219} \\
\multicolumn{1}{|c|}{} &
\multicolumn{1}{|c|}{} &
  Ours &
  \textbf{12.88} &
  \textbf{14.71} &
  15.40 &
  \textbf{16.13} &
  \textbf{0.216} &
  0.297 &
  0.343 &
  0.391 &
  0.462 &
  0.383 &
  0.366 &
  0.352 \\
\multicolumn{1}{|c|}{} &
\multicolumn{1}{|c|}{} &
  Oracle &
  14.89 &
  16.45 &
  17.42 &
  18.45 &
  0.365 &
  0.471 &
  0.525 &
  0.576 &
  0.303 &
  0.237 &
  0.200 &
  0.174 \\ \cline{2-15}
\multicolumn{1}{|c|}{} &
\multicolumn{1}{|c|}{\multirow{3}{*}{Room}} &
  3DGS &
  10.18 &
  11.51 &
  11.59 &
  12.21 &
  0.404 &
  0.494 &
  0.510 &
  0.552 &
  0.606 &
  0.559 &
  0.556 &
  0.515 \\
\multicolumn{1}{|c|}{} &
\multicolumn{1}{|c|}{} &
  Ours &
  \textbf{17.21} &
  \textbf{18.11} &
  \textbf{18.87} &
  \textbf{19.63} &
  \textbf{0.668} &
  \textbf{0.719} &
  \textbf{0.732} &
  \textbf{0.757} &
  \textbf{0.352} &
  \textbf{0.360} &
  \textbf{0.326} &
  \textbf{0.295} \\
\multicolumn{1}{|c|}{} &
\multicolumn{1}{|c|}{} &
  Oracle &
  20.66 &
  22.31 &
  23.80 &
  24.59 &
  0.758 &
  0.801 &
  0.839 &
  0.864 &
  0.217 &
  0.188 &
  0.160 &
  0.156 \\ \cline{2-15}
\multicolumn{1}{|c|}{} &
\multicolumn{1}{|c|}{\multirow{3}{*}{Trex}} &
  3DGS &
  10.72 &
  11.72 &
  13.11 &
  14.14 &
  0.322 &
  0.417 &
  0.492 &
  0.548 &
  0.520 &
  0.446 &
  0.394 &
  0.351 \\
\multicolumn{1}{|c|}{} &
\multicolumn{1}{|c|}{} &
  Ours &
  \textbf{14.90} &
  \textbf{15.90} &
  \textbf{16.75} &
  \textbf{17.37} &
  \textbf{0.480} &
  \textbf{0.537} &
  \textbf{0.567} &
  \textbf{0.625} &
  \textbf{0.358} &
  \textbf{0.362} &
  \textbf{0.348} &
  \textbf{0.305} \\
\multicolumn{1}{|c|}{} &
\multicolumn{1}{|c|}{} &
  Oracle &
  17.76 &
  19.58 &
  20.84 &
  22.83 &
  0.591 &
  0.669 &
  0.714 &
  0.786 &
  0.284 &
  0.226 &
  0.192 &
  0.134 \\ \hline
\multicolumn{2}{|c|}{\multirow{3}{*}{Mean}} &
  3DGS &
  \cellcolor{tabthird}12.25 &
  \cellcolor{tabthird}13.75 &
  \cellcolor{tabthird}15.26 &
  \cellcolor{tabthird}16.17 &
  \cellcolor{tabthird}0.306 &
  \cellcolor{tabthird}0.407 &
  \cellcolor{tabthird}0.485 &
  \cellcolor{tabthird}0.533 &
  \cellcolor{tabthird}0.471 &
  \cellcolor{tabthird}0.388 &
  \cellcolor{tabthird}0.334 &
  \cellcolor{tabthird}0.299 \\
\multicolumn{2}{|c|}{} &
  Ours &
  \cellcolor{tabfirst}\textbf{15.94} &
  \cellcolor{tabfirst}\textbf{17.17} &
  \cellcolor{tabfirst}\textbf{18.15} &
  \cellcolor{tabfirst}\textbf{18.74} &
  \cellcolor{tabfirst}\textbf{0.439} &
  \cellcolor{tabfirst}\textbf{0.497} &
  \cellcolor{tabfirst}\textbf{0.541} &
  \cellcolor{tabfirst}\textbf{0.571} &
  \cellcolor{tabfirst}\textbf{0.365} &
  \cellcolor{tabfirst}\textbf{0.337} &
  \cellcolor{tabfirst}\textbf{0.309} &
  \cellcolor{tabfirst}\textbf{0.288} \\
\multicolumn{2}{|c|}{} &
  Oracle &
  18.29 &
  19.95 &
  21.02 &
  22.05 &
  0.539 &
  0.622 &
  0.660 &
  0.701 & 
  0.266 &
  0.213 &
  0.188 &
  0.167 \\ \hline
\end{tabular}}
\vspace{-0.8 em}
\caption{Quantitative results in NeRF-LLFF~\cite{mildenhall2019llff} dataset. The best performance except oracle is \textbf{bolded}.}
\label{tab:result_llff}
\vspace{-0.8 em}
\end{table*}

\subsection{Unsupervised Smoothness Constraint} \label{sec:3_DepthSmooth}
Even though each independently estimated depth was fitted to the COLMAP points, conflicts often arise.
We introduce an unsupervised constraint for geometry smoothness inspired by~\cite{godard2017unsupervised} to regularize the conflict.
This constraint implies that points in similar 3D positions have similar depths on the image plane.
We utilize the Canny edge detector~\cite{canny1986computational} as a mask to ensure that it does not regularize the area with significant differences in depth along the boundaries.
For a depth $d_i$ and its adjacent depth $d_j$, we regularize the difference between them:
\begin{align}
    \mathcal{L}_{smooth} = \sum_{d_j\in \text{adj}(d_i)} \mathbbm{1}_{ne}{(d_i,d_j)}{\cdot} \begin{Vmatrix} d_i - d_j \end{Vmatrix}^2
    \label{eq:smoothloss}
\end{align}
where $\mathbbm{1}_{ne}$ is a indicator function that signifies whether both depths are \textit{not} in edge.

We conclude the final loss terms by incorporating the depth loss from \eqnref{depthloss} and smoothness loss and smoothness loss from \eqnref{smoothloss} with their own hyperparameters $\lambda_{depth}$ and $\lambda_{smooth}$:
\begin{equation}
\begin{aligned}
    \mathcal{L} = (1{-}\lambda_{ssim})\mathcal{L}_{color} &+ \lambda_{ssim}\mathcal{L}_{D-SSIM} \\ 
    + \lambda_{depth}\mathcal{L}_{depth} &+ \lambda_{smooth}\mathcal{L}_{smooth}
    \label{eq:finalloss}
\end{aligned}
\end{equation}
where the preceding two loss terms $\mathcal{L}_{color}, \mathcal{L}_{D-SSIM}$ correspond to the original 3D Gaussian splatting losses.~\cite{kerbl20233d}

\subsection{Modification for Few-Shot Learning} \label{sec:3_OptSet}
We modify two optimization techniques from the original paper to create 3D scenes with limited images.
The techniques employed in 3DGS were designed under the assumption of utilizing a substantial number of images, potentially hindering convergence in a few-shot setting. 
Through iterative experiments, we confirm this and modify the techniques to suit the few-shot setting.
Firstly, we set the maximum degree of spherical harmonics (SH) to 1.
This prevents overfitting of spherical harmonic coefficients responsible for high frequencies due to insufficient information.
Secondly, we implement an early-stop policy based on depth loss. 
We configure \eqnref{finalloss} to be primarily driven by color loss, while employing the depth loss and the smoothness loss as guiding factors. 
Hence, overfitting gradually emerges due to the predominant influence of color loss.
We use a moving averaged depth loss to halt optimization when the splats start to deviate from the depth guide.
Lastly, we remove the periodic reset process.
We observe that resetting the opacity $\alpha$ of all splats leads to irreversible and detrimental consequences. 
Due to a lack of information from the limited images, the inability to restore the opacity of splats led to scenarios where either all splats were removed or trapped in local optima, causing unexpected outcomes and optimization failures. 
As a result of the aforementioned techniques, we achieve stable optimization in few-shot learning.
\section{Experiment} \label{sec:experiment}

\subsection{Experiment settings}
\paragraph{Datasets.}
We evaluate our method on NeRF-LLFF~\cite{mildenhall2019llff} dataset. 
NeRF-LLFF includes 8 scenes with forward-facing cameras, and we split the images of each scene into train and test sets.
We use the image outer edge of the camera group as the train set based on the convex hull algorithm~\cite{preparata2012computational} due to the forward-facing camera distribution.
For each experiment, we optimize the scene with k-shot (k=2,3,4,5) images randomly selected from the train set and evaluate on the same test set.
We use ten randomly selected seeds and report the average of ten experiments.

\begin{figure*}[t]
    \centering
    \begin{minipage}[]{0.03\linewidth}%
        \scalebox{0.9}{
            \rotatebox{90}{
                \begin{tabular}[t]{>{\arraybackslash}p{1cm}
                                   >{\arraybackslash}p{4.5cm}
                                   >{\arraybackslash}p{4.5cm}
                                   >{\arraybackslash}p{4.5cm}
                                   >{\arraybackslash}p{3.3cm}
                                   }
                     & (iv) Horns
                     & (iii) Fortress
                     & (ii) Room 
                     & (i) Fern
                \end{tabular}
            }
        }
        \end{minipage}
    \begin{minipage}[]{0.03\linewidth}%
        \scalebox{0.9}{
            \rotatebox{90}{
            \begin{tabular}[t]{>{\centering\arraybackslash}p{0.2cm}
                               >{\centering\arraybackslash}p{1.9cm}
                               >{\centering\arraybackslash}p{2.9cm}
                               >{\centering\arraybackslash}p{1.9cm}
                               >{\centering\arraybackslash}p{2.9cm}
                               >{\centering\arraybackslash}p{1.9cm}
                               >{\centering\arraybackslash}p{2.9cm}
                               >{\centering\arraybackslash}p{1.9cm}
                               >{\centering\arraybackslash}p{2cm}
                               }
                 & 5-view & 2-view              
                 & 5-view & 2-view
                 & 5-view & 2-view
                 & 5-view & 2-view 
            \end{tabular}
            }
        }
        \end{minipage}
    \begin{minipage}[]{0.93\linewidth}
        \centering
        \renewcommand{\wp}{0.197 \linewidth}
        \addtocounter{subfigure}{-5}
        \subfloat{\includegraphics[width=\wp]{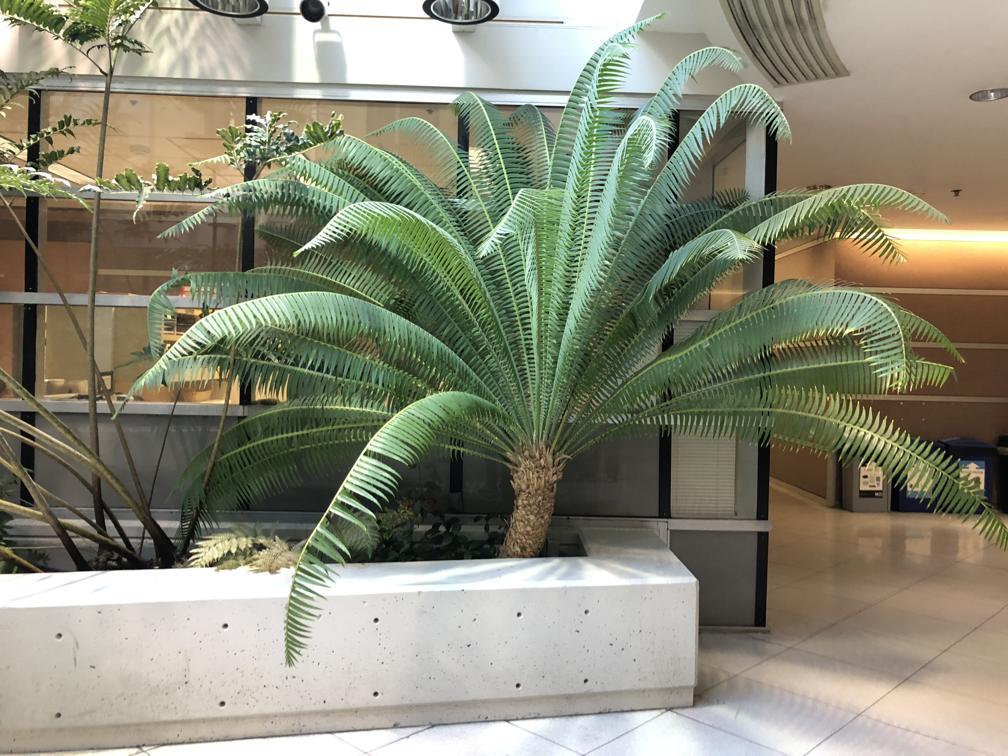}}
        \hfill
        \subfloat{\includegraphics[width=\wp]{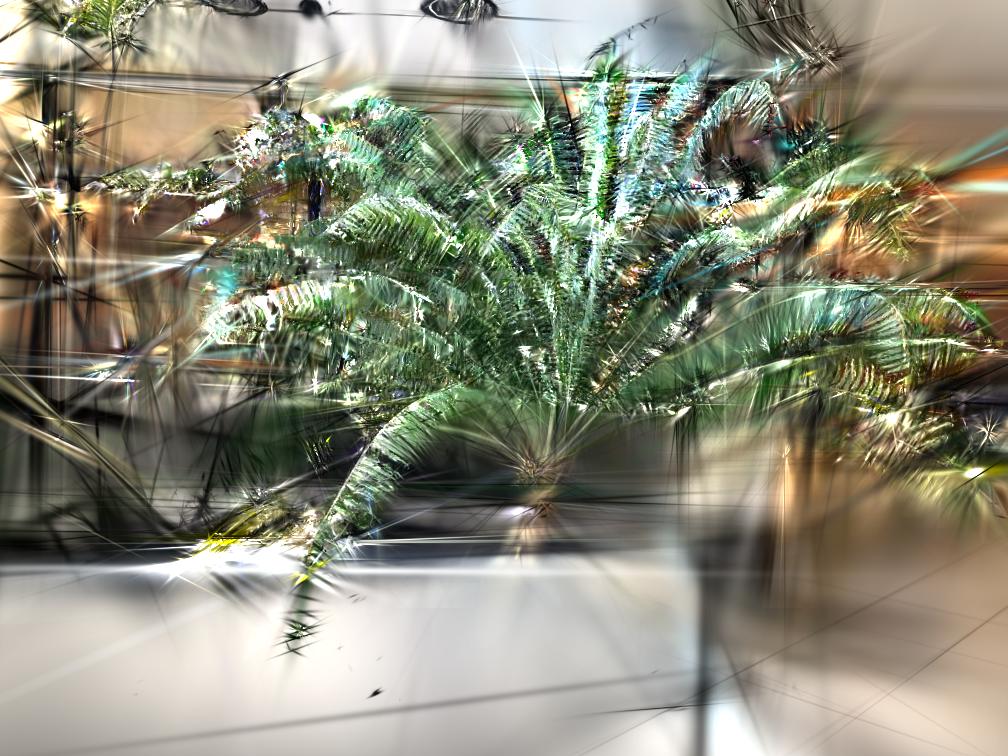}}
        \hfill
        \subfloat{\includegraphics[width=\wp]{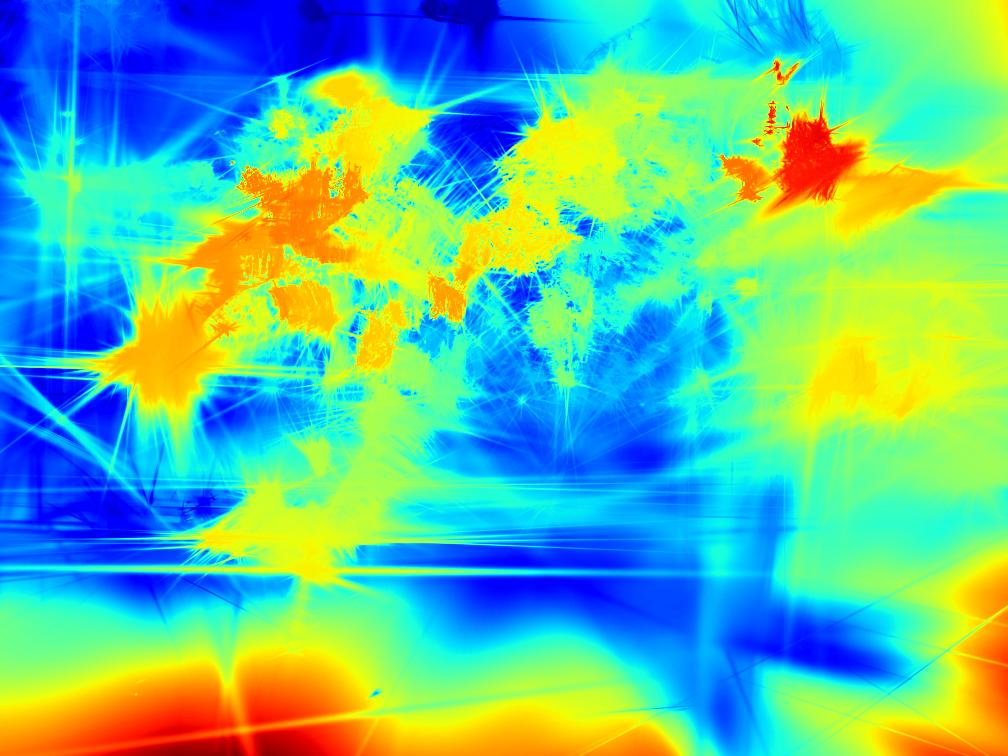}}
        \hfill
        \subfloat{\includegraphics[width=\wp]{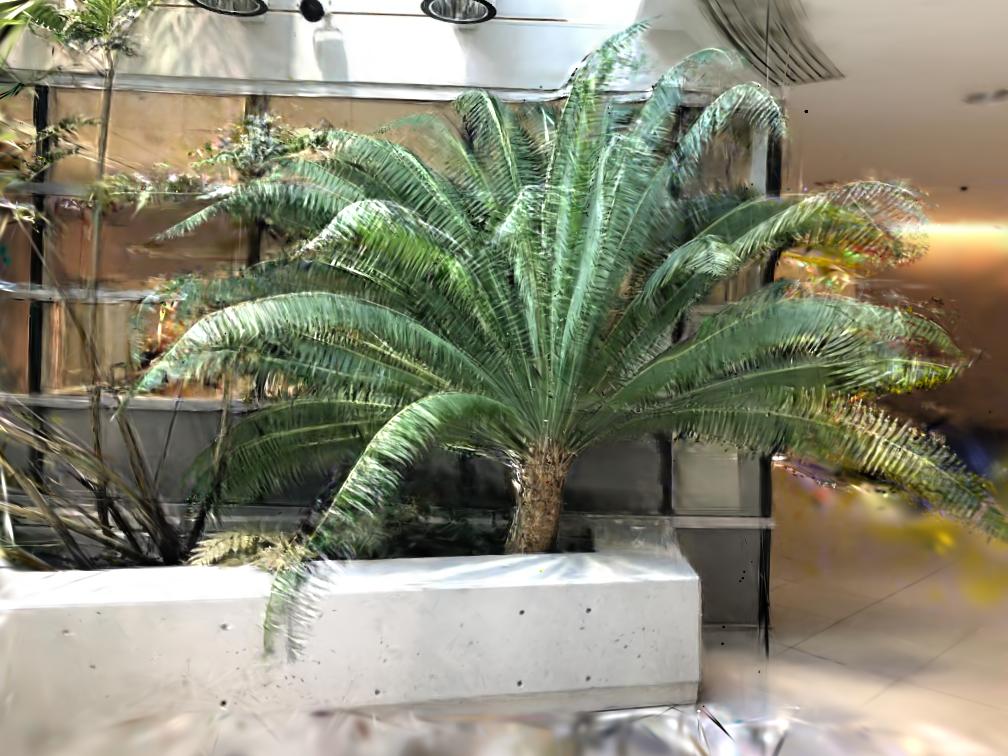}}
        \hfill
        \subfloat{\includegraphics[width=\wp]{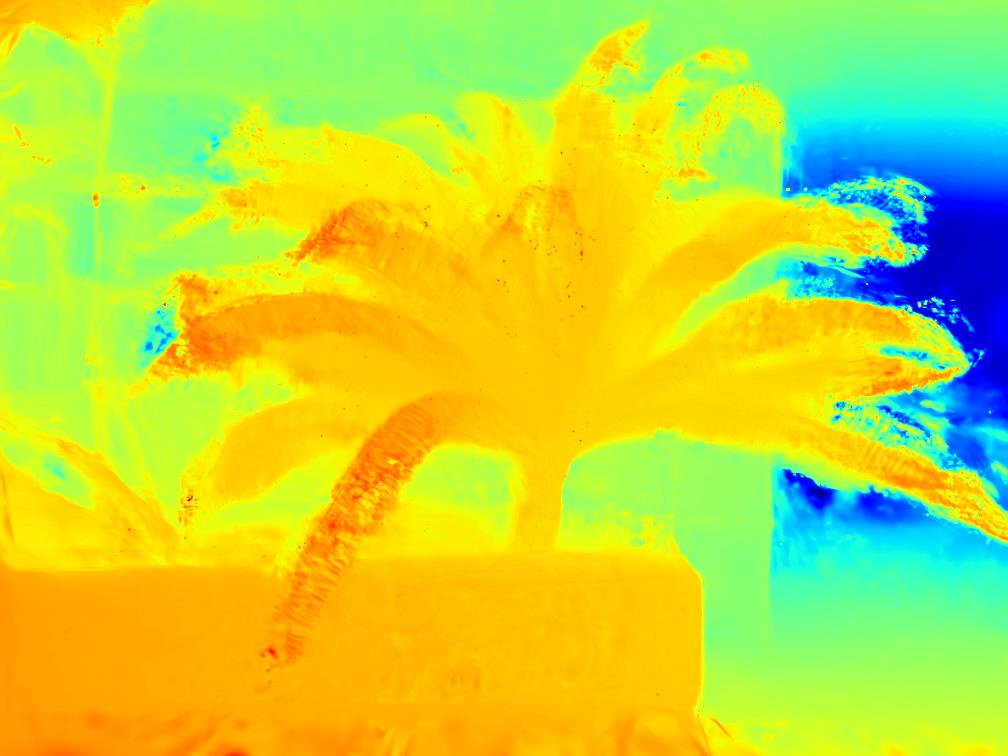}}
        \\
        \vspace{+0.05cm}
        \addtocounter{subfigure}{-5}
        \subfloat{\includegraphics[width=\wp]{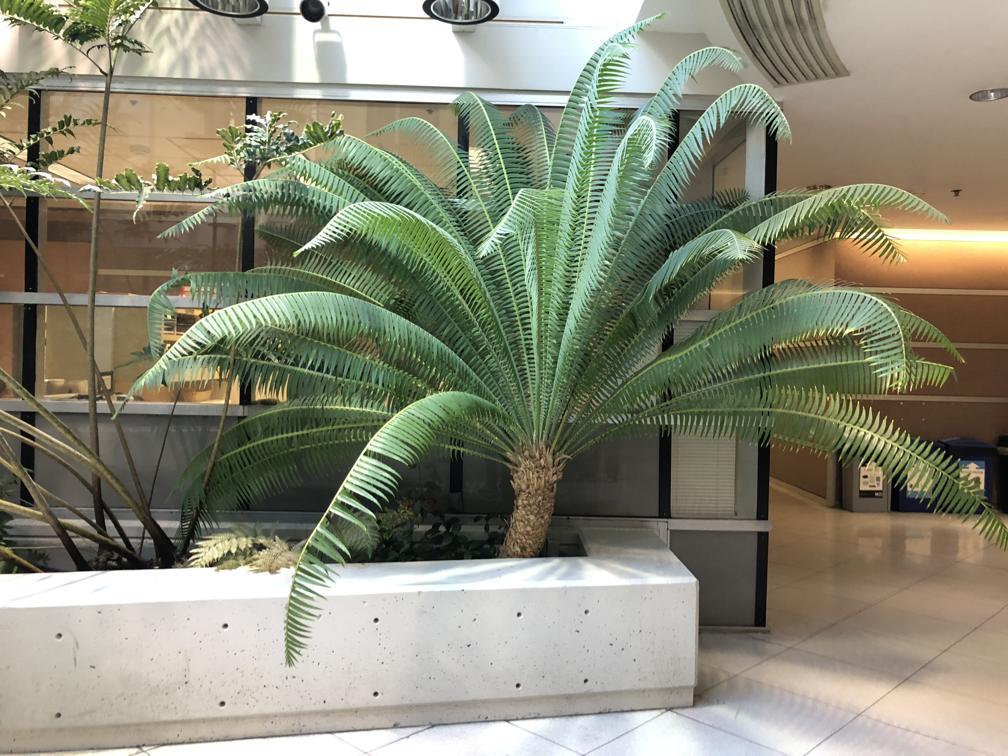}}
        \hfill
        \subfloat{\includegraphics[width=\wp]{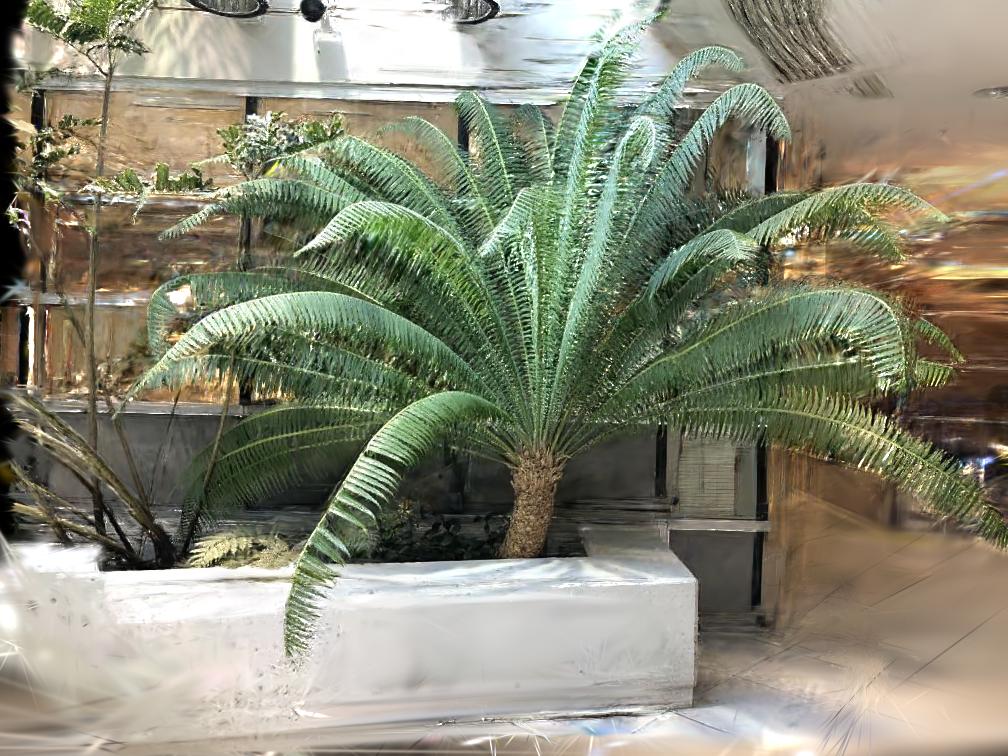}}
        \hfill
        \subfloat{\includegraphics[width=\wp]{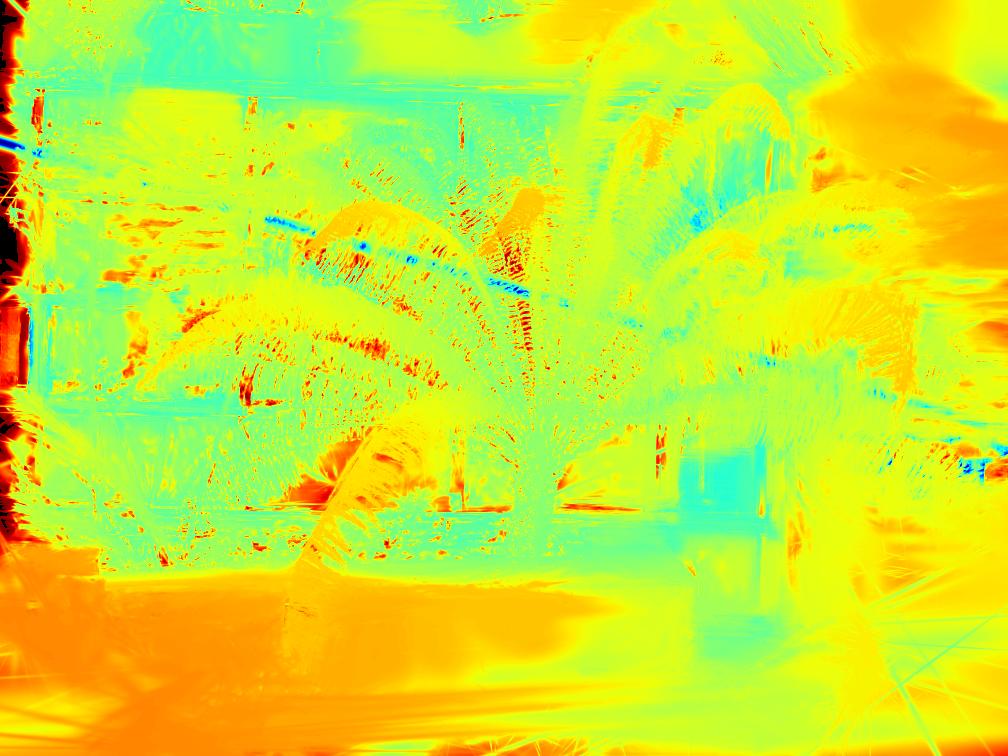}}
        \hfill
        \subfloat{\includegraphics[width=\wp]{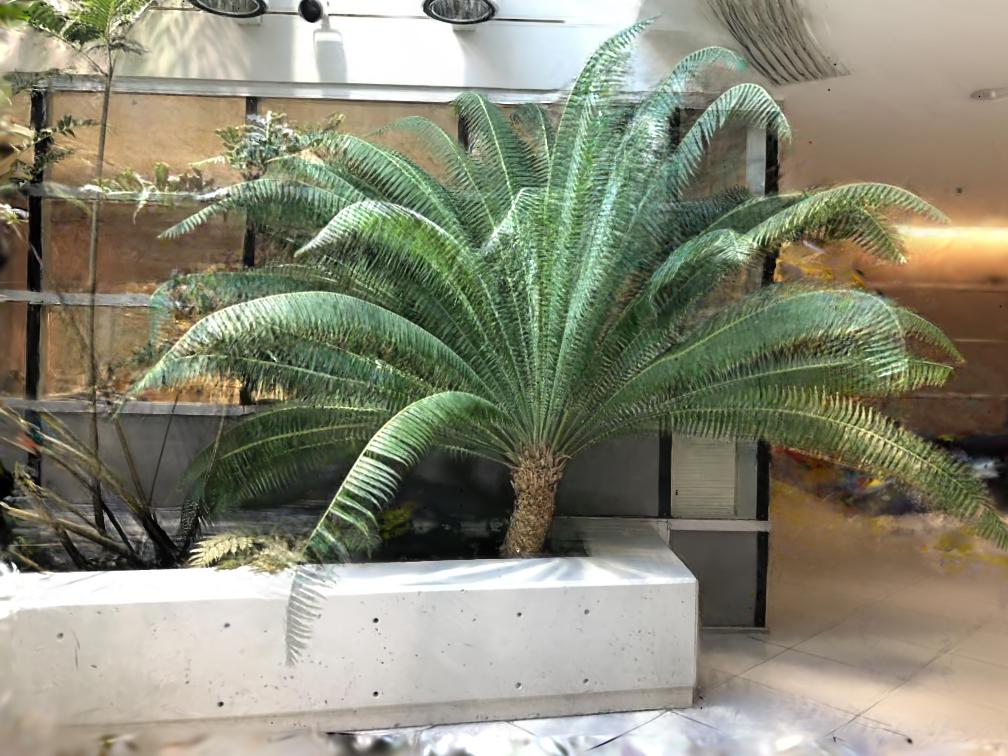}}
        \hfill
        \subfloat{\includegraphics[width=\wp]{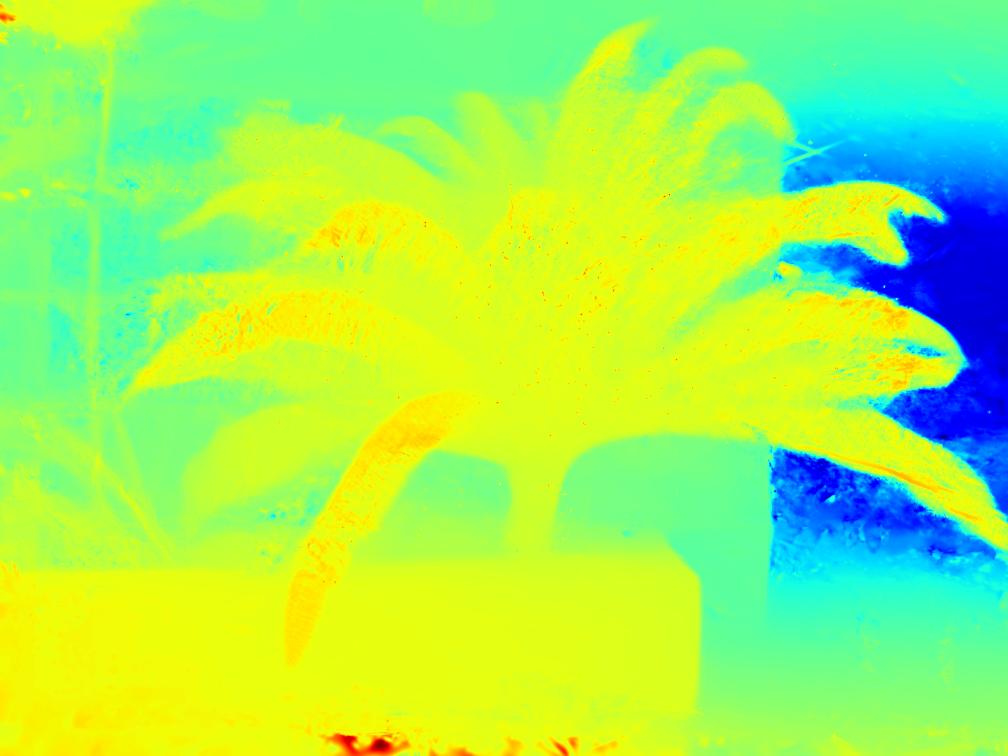}}
        \\
        \vspace{+0.2cm}
        \addtocounter{subfigure}{-5}
        \subfloat{\includegraphics[width=\wp]{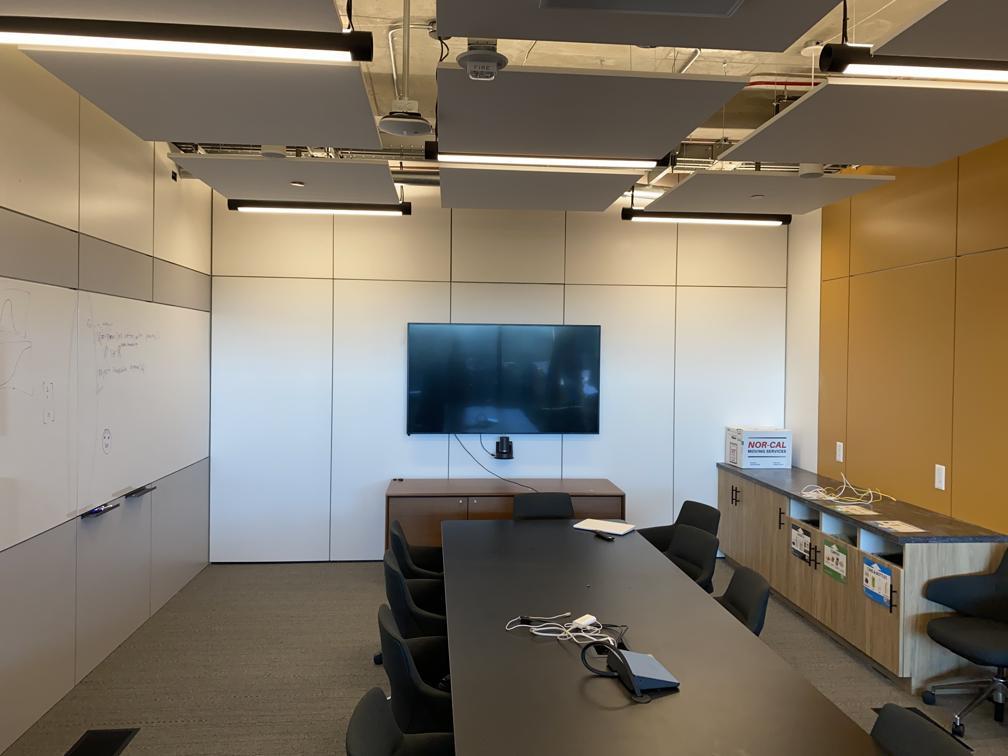}}
        \hfill
        \subfloat{\includegraphics[width=\wp]{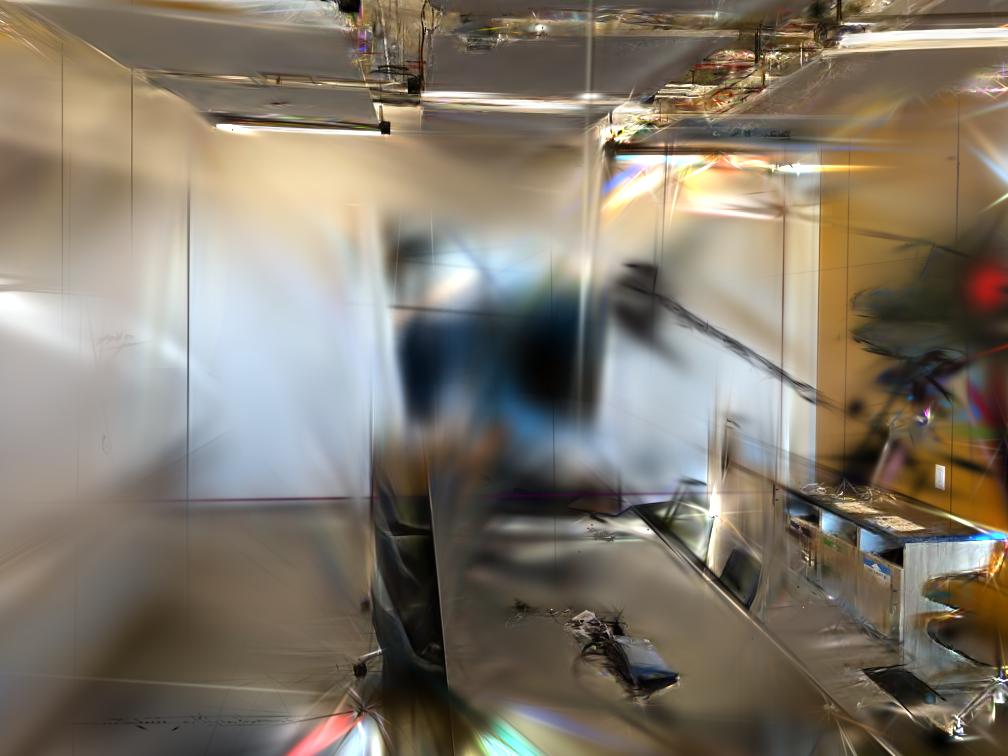}}
        \hfill
        \subfloat{\includegraphics[width=\wp]{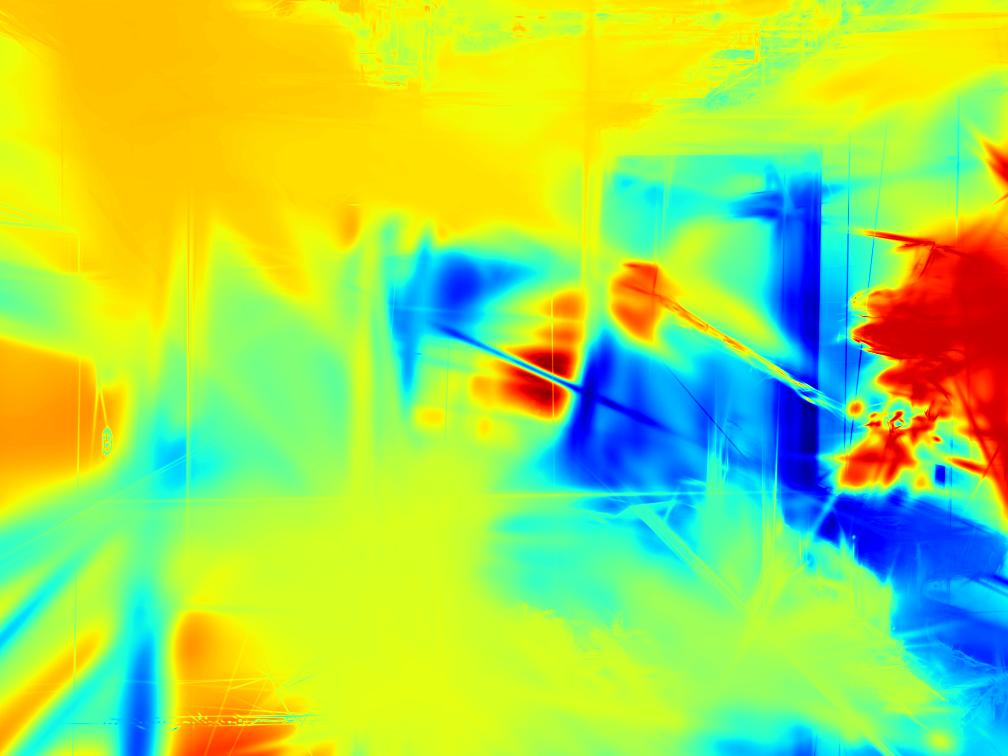}}
        \hfill
        \subfloat{\includegraphics[width=\wp]{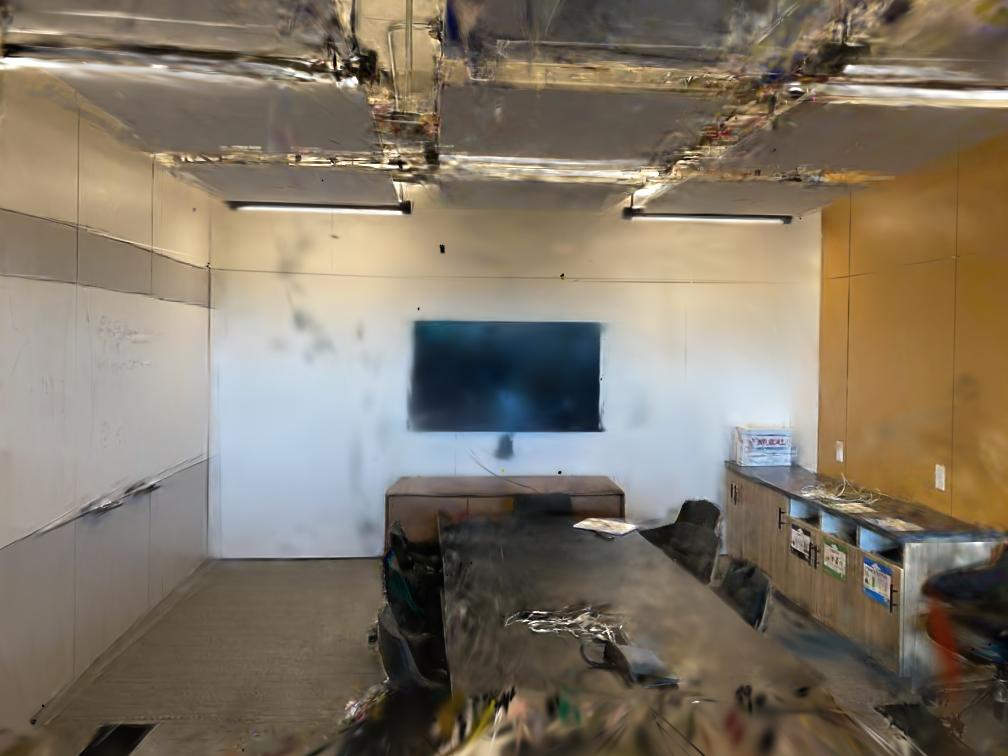}}
        \hfill
        \subfloat{\includegraphics[width=\wp]{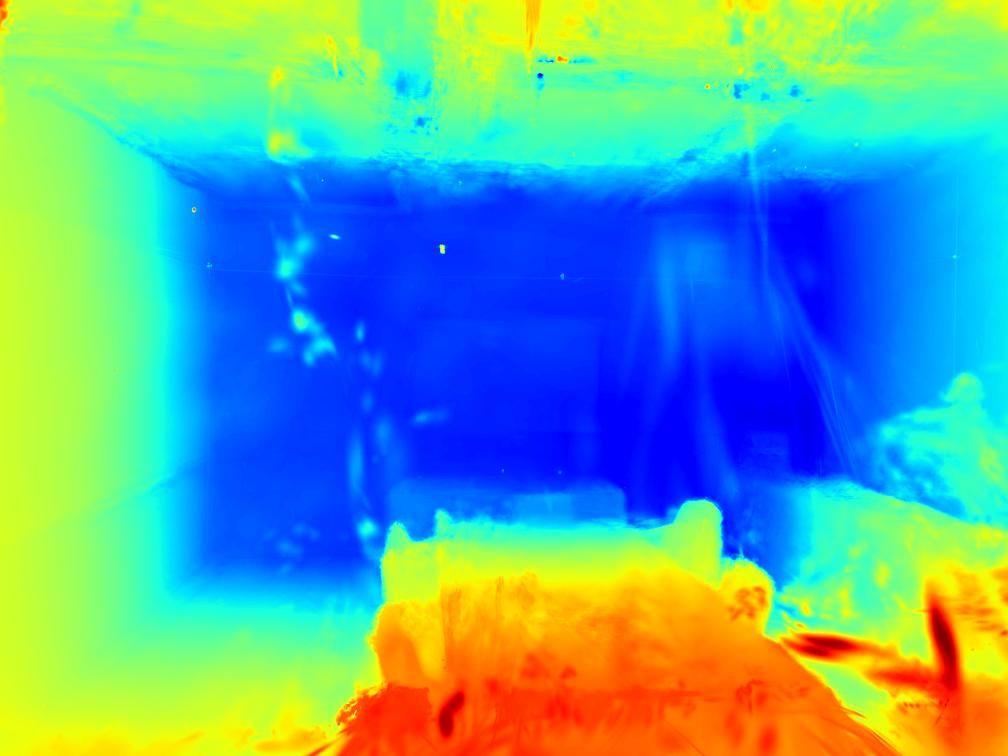}}
        \\
        \vspace{+0.05cm}
        \addtocounter{subfigure}{-5}
        \subfloat{\includegraphics[width=\wp]{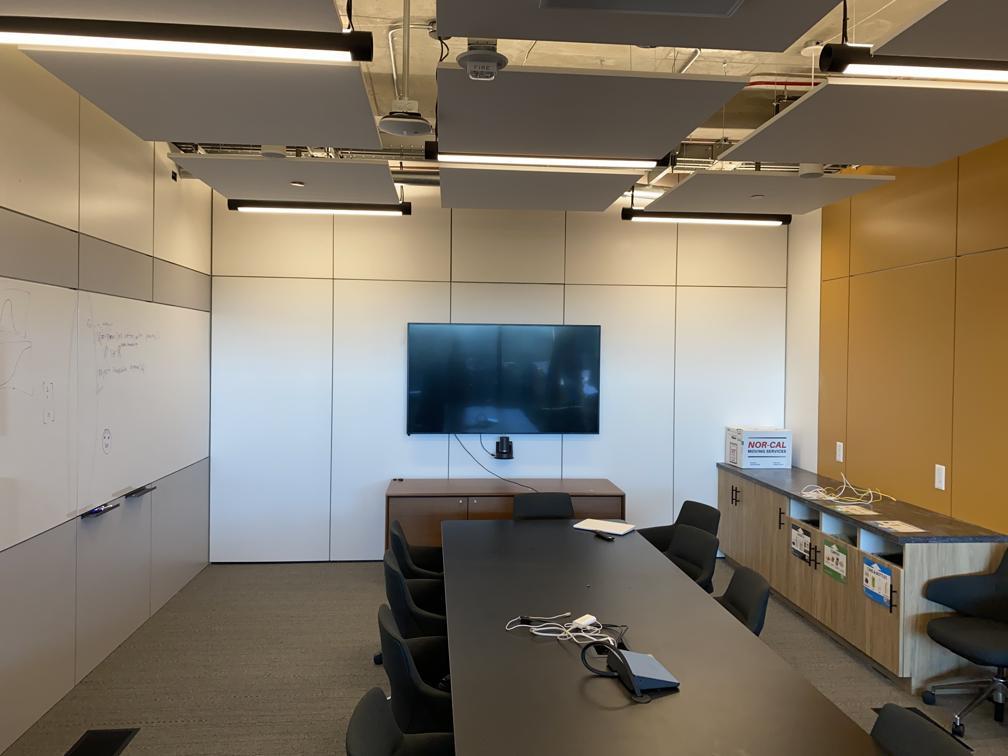}}
        \hfill
        \subfloat{\includegraphics[width=\wp]{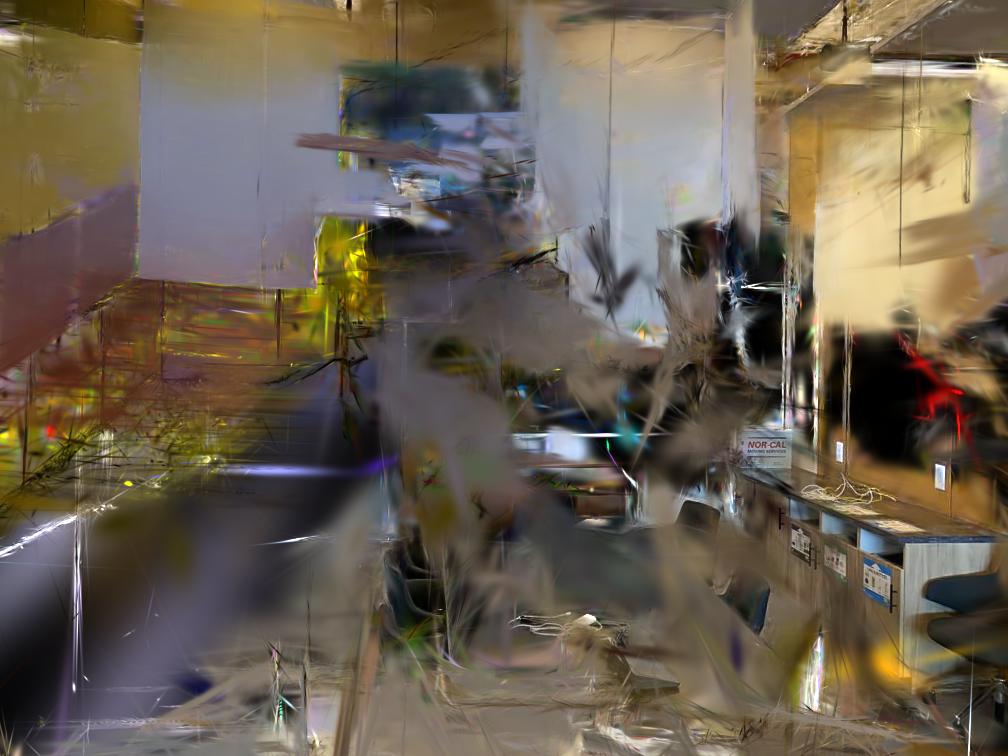}}
        \hfill
        \subfloat{\includegraphics[width=\wp]{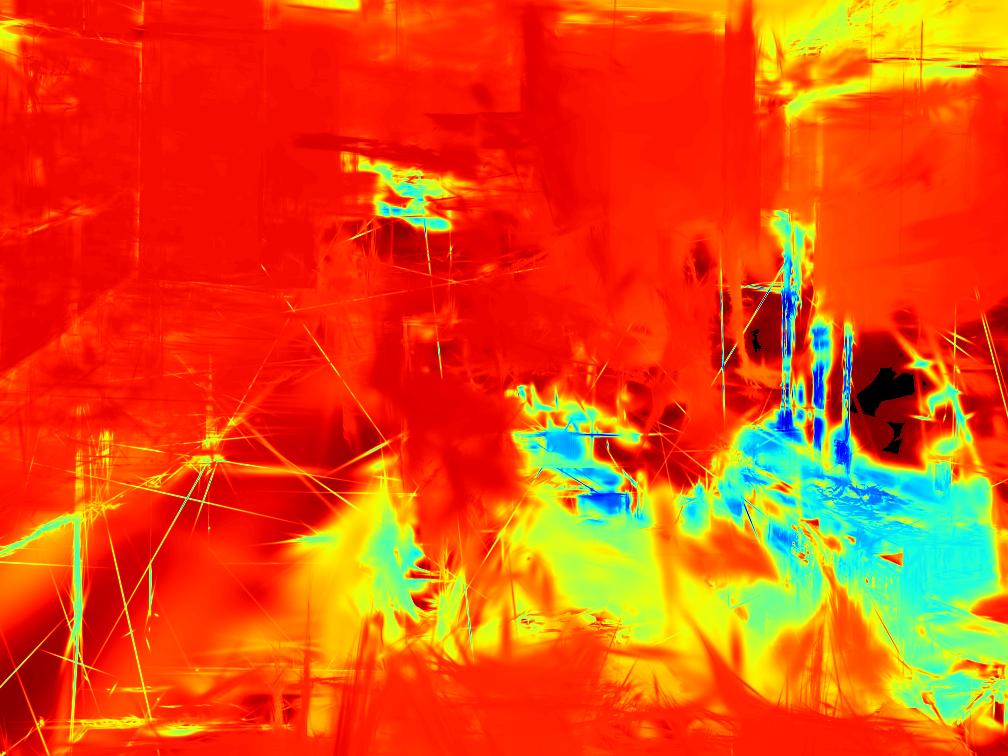}}
        \hfill
        \subfloat{\includegraphics[width=\wp]{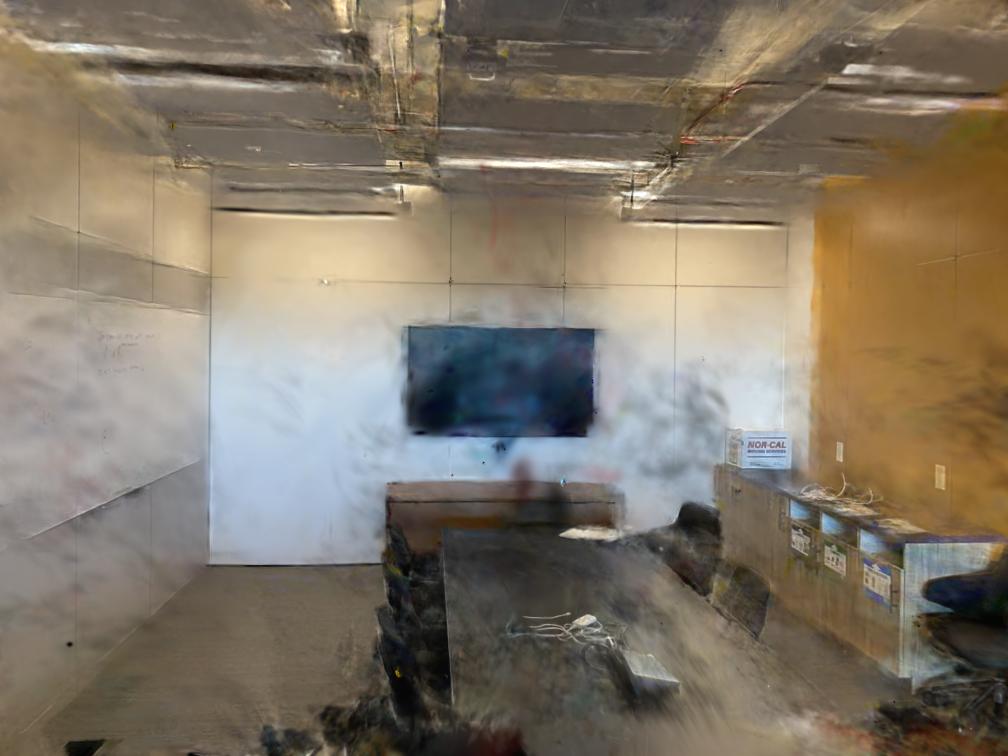}}
        \hfill
        \subfloat{\includegraphics[width=\wp]{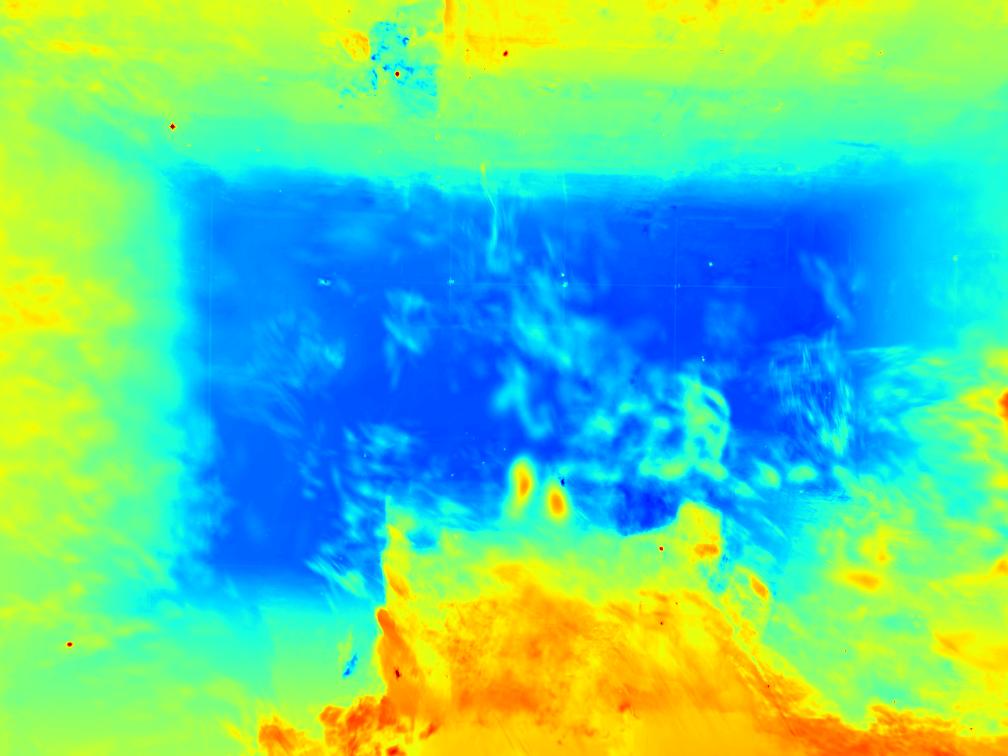}}
        \\
        \vspace{+0.2cm}
        \addtocounter{subfigure}{-5}
        \subfloat{\includegraphics[width=\wp]{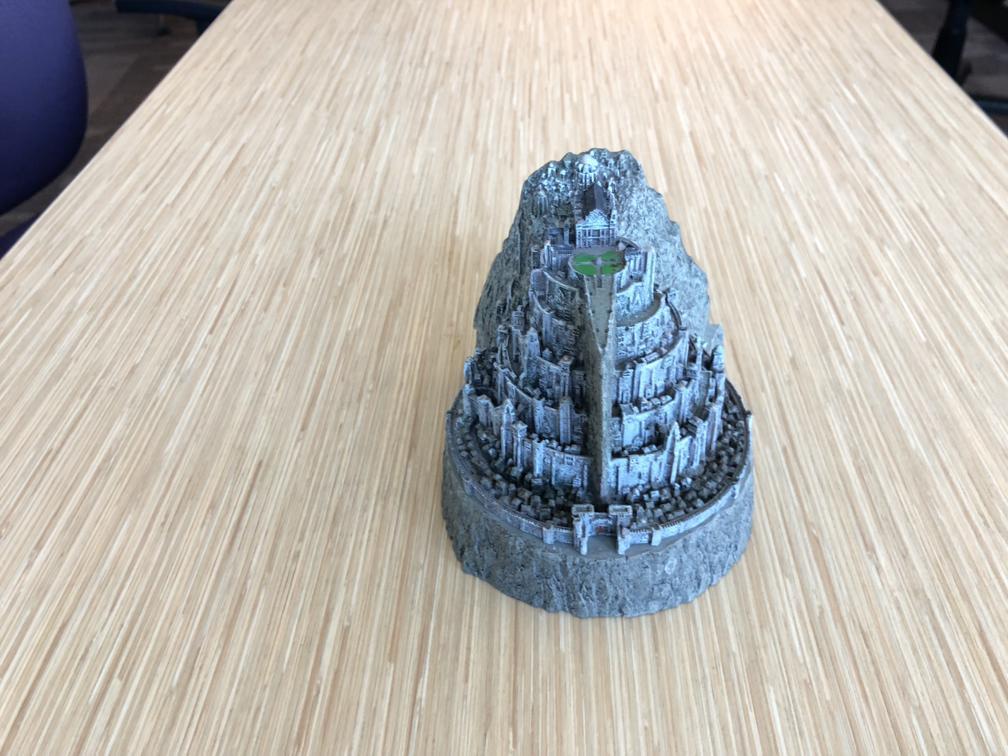}}
        \hfill
        \subfloat{\includegraphics[width=\wp]{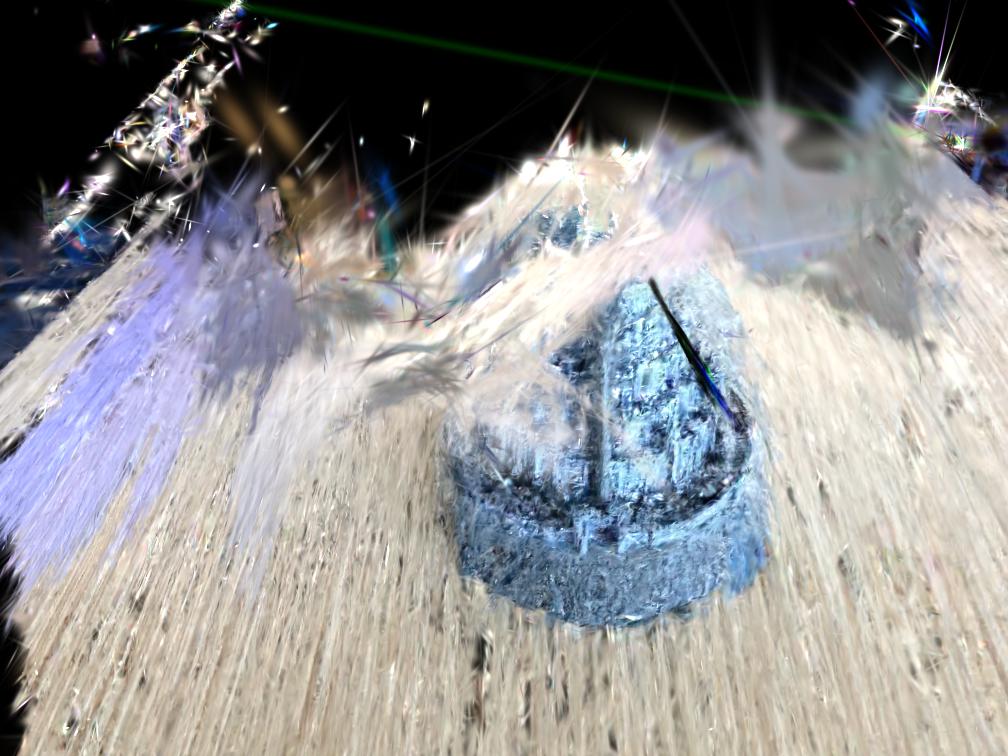}}
        \hfill
        \subfloat{\includegraphics[width=\wp]{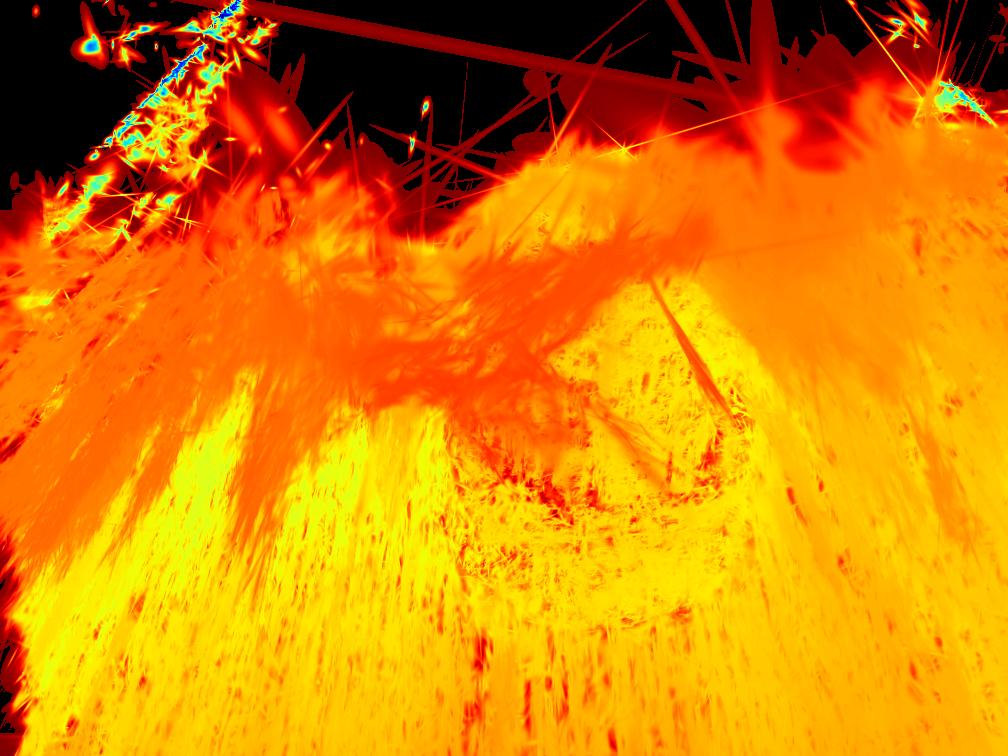}}
        \hfill
        \subfloat{\includegraphics[width=\wp]{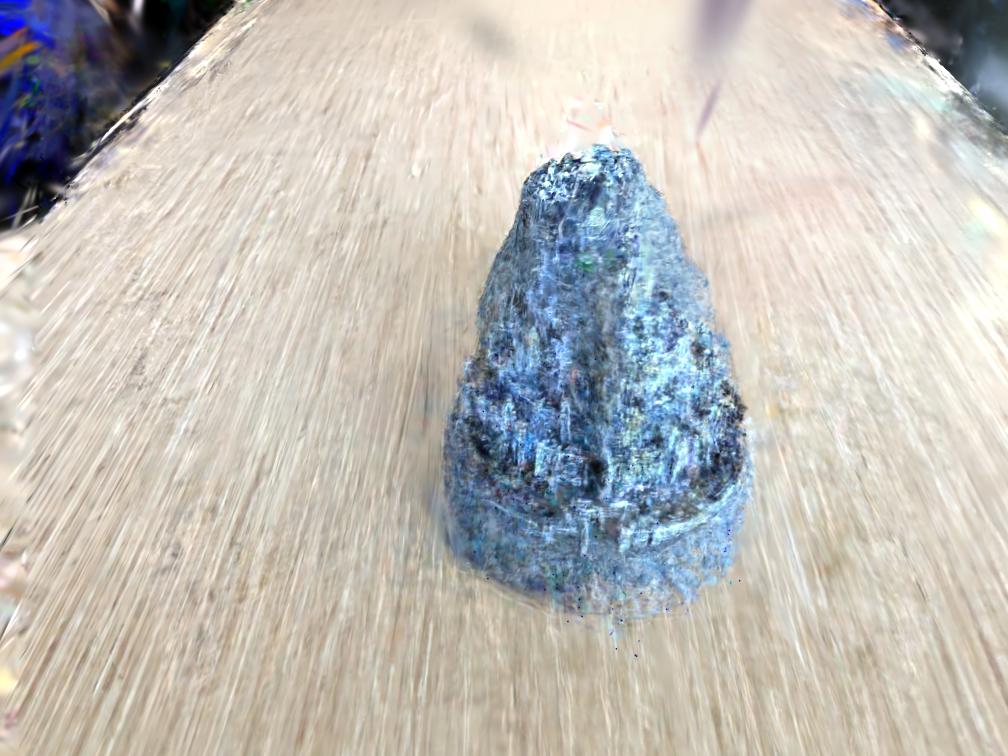}}
        \hfill
        \subfloat{\includegraphics[width=\wp]{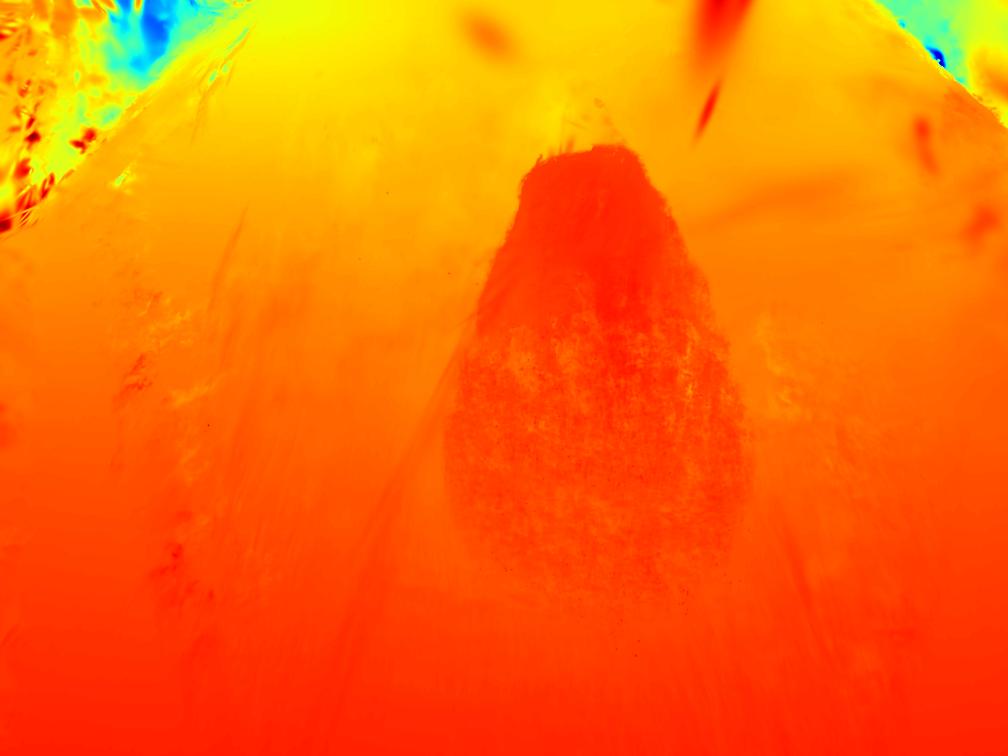}}
        \\
        \vspace{+0.05cm}
        \addtocounter{subfigure}{-5}
        \subfloat{\includegraphics[width=\wp]{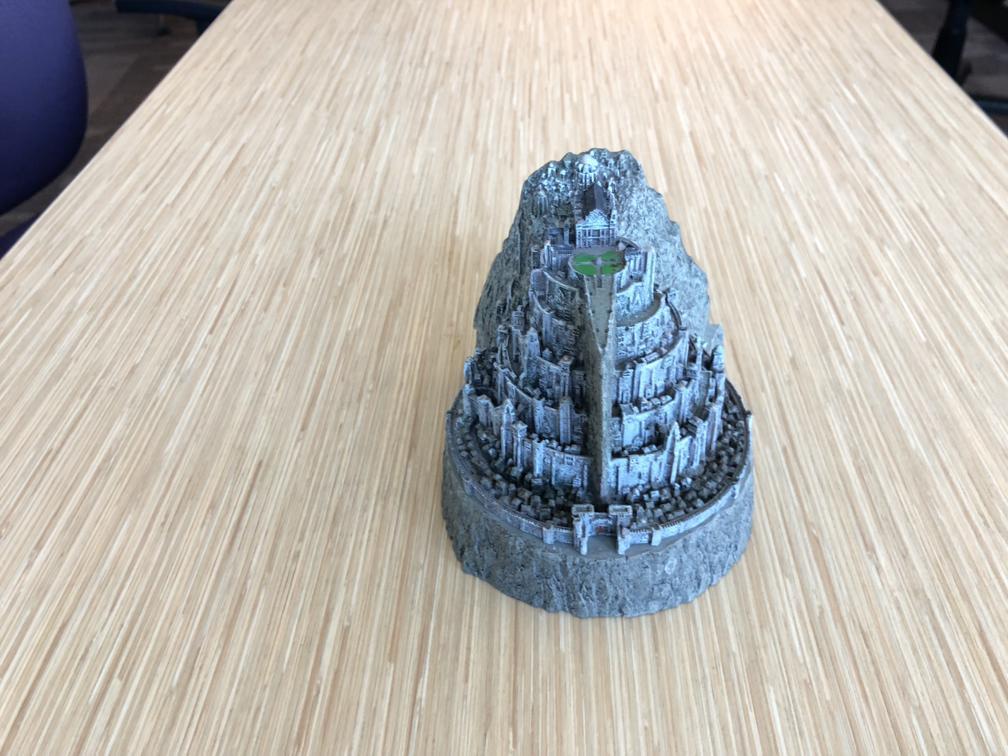}}
        \hfill
        \subfloat{\includegraphics[width=\wp]{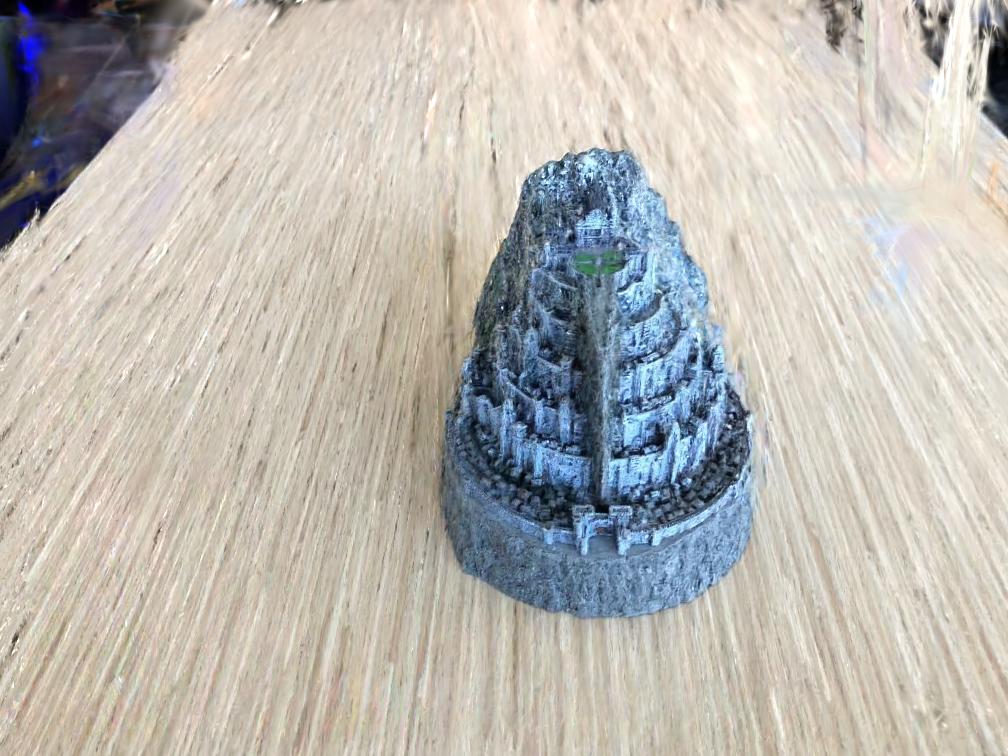}}
        \hfill
        \subfloat{\includegraphics[width=\wp]{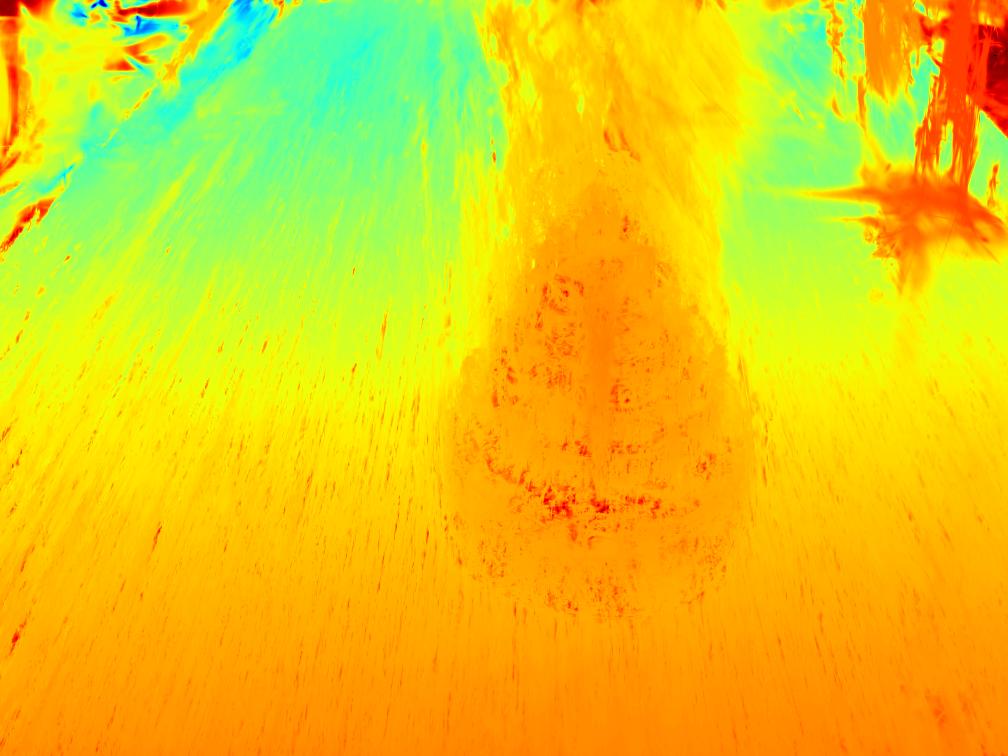}}
        \hfill
        \subfloat{\includegraphics[width=\wp]{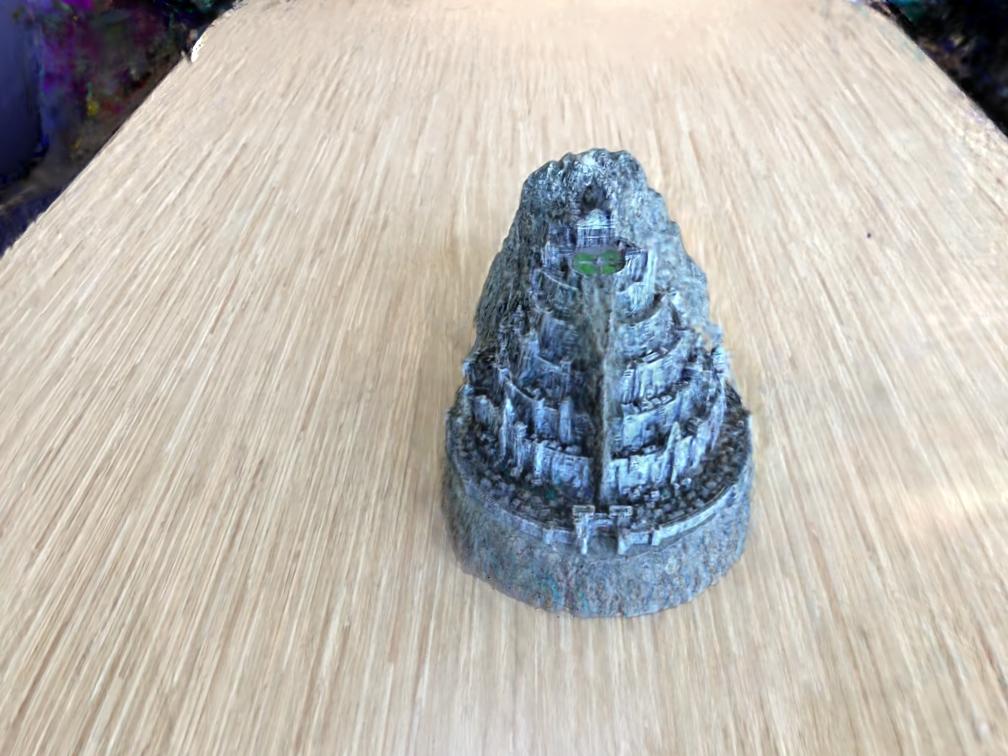}}
        \hfill
        \subfloat{\includegraphics[width=\wp]{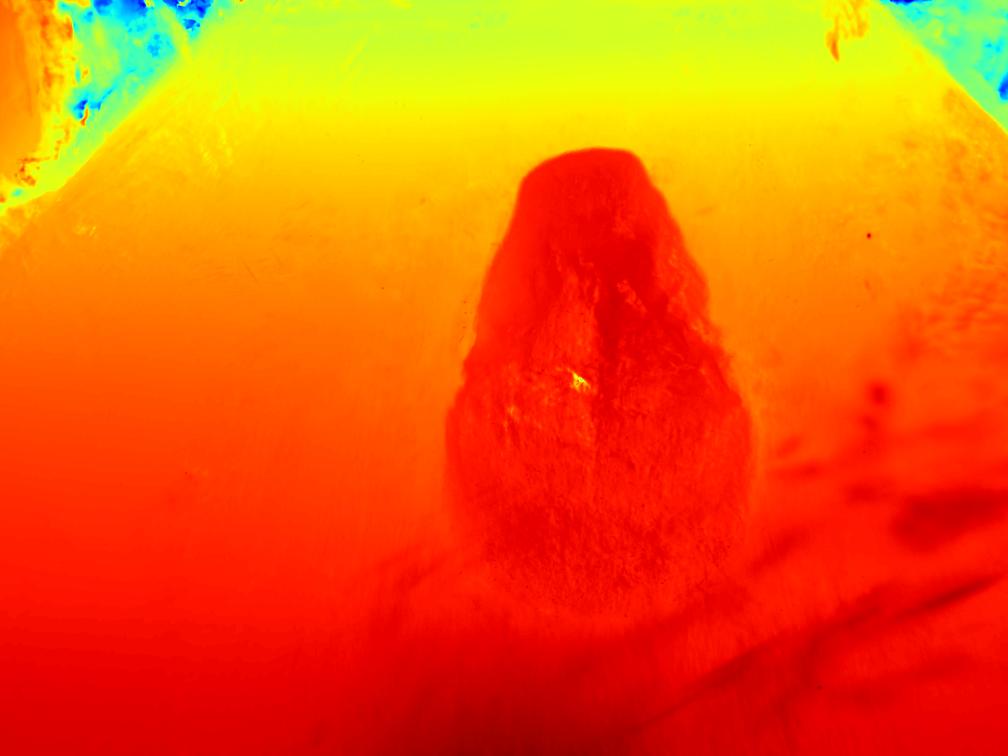}}
        \\
        \vspace{+0.2cm}
        \addtocounter{subfigure}{-5}
        \subfloat{\includegraphics[width=\wp]{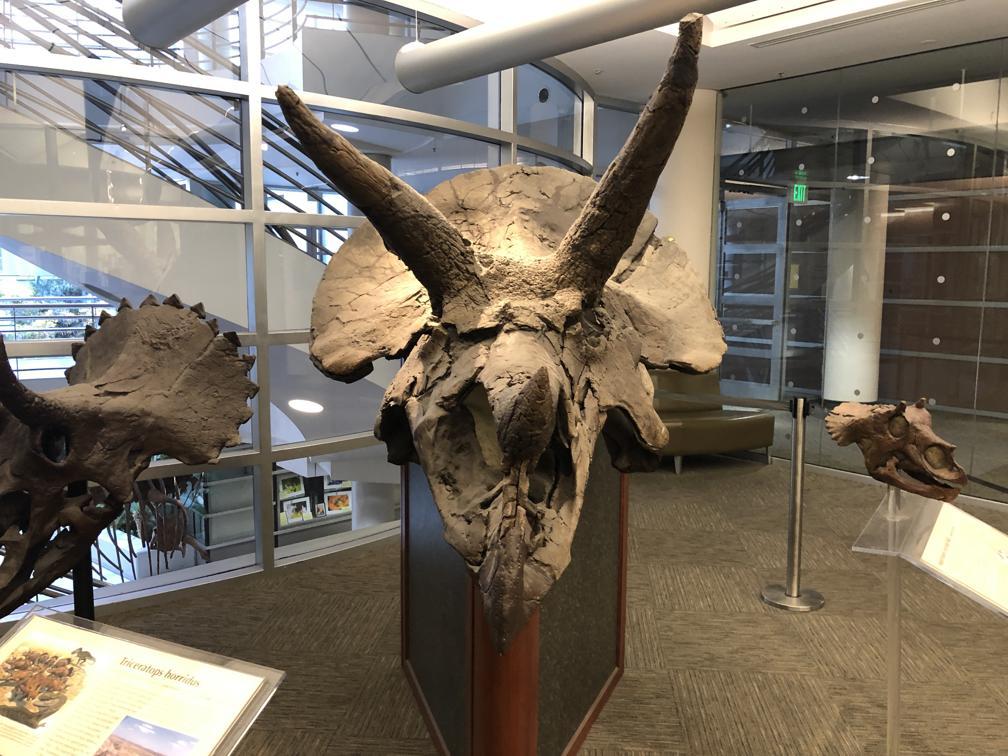}}
        \hfill
        \subfloat{\includegraphics[width=\wp]{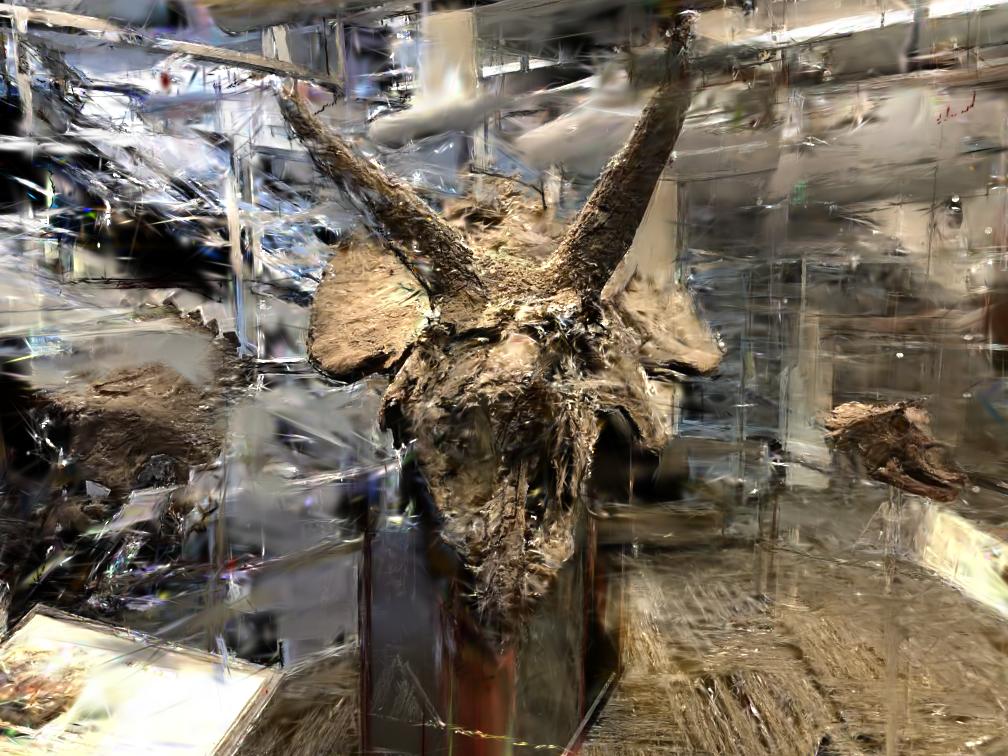}}
        \hfill
        \subfloat{\includegraphics[width=\wp]{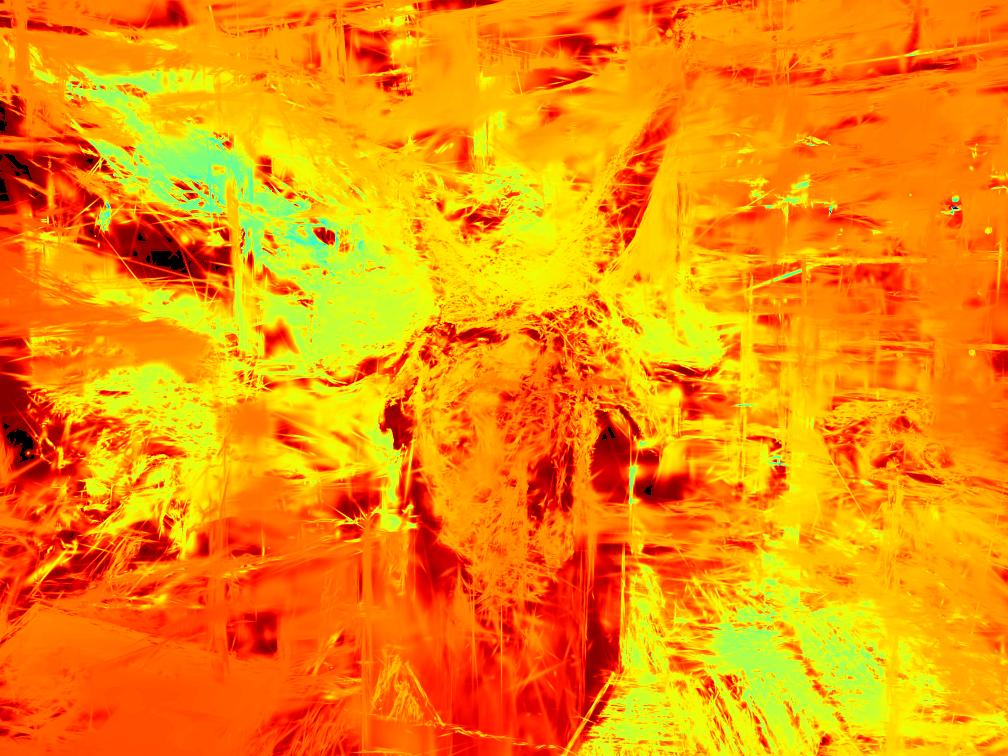}}
        \hfill
        \subfloat{\includegraphics[width=\wp]{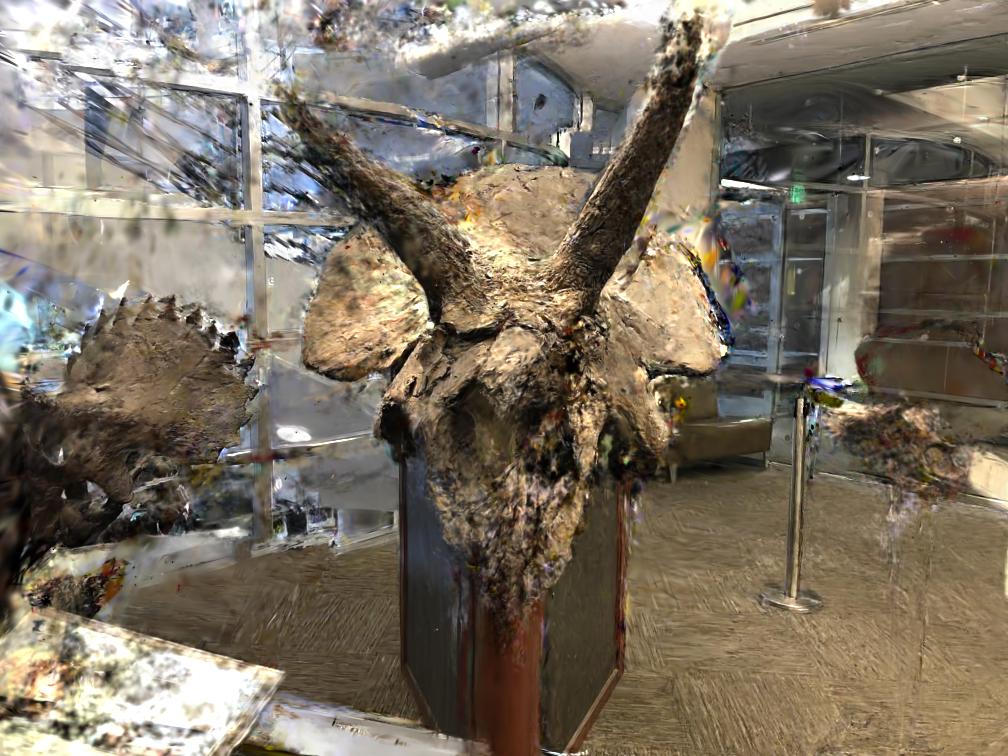}}
        \hfill
        \subfloat{\includegraphics[width=\wp]{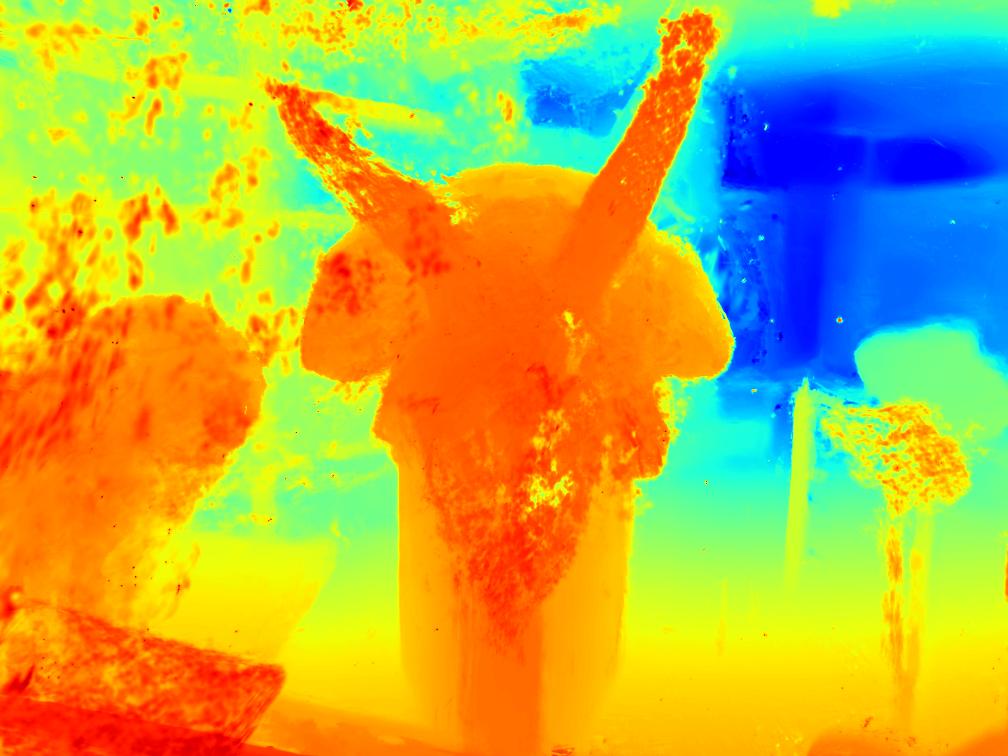}}
        \\
        \vspace{+0.05cm}
        \addtocounter{subfigure}{-5}
        \subfloat[GT]{\includegraphics[width=\wp]{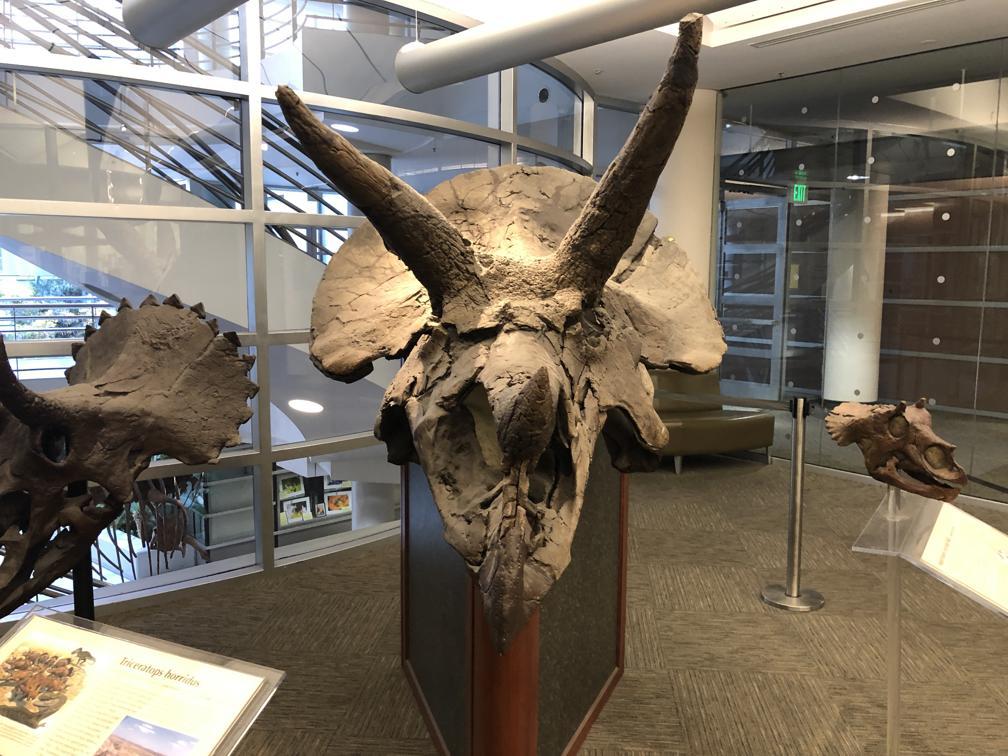}}
        \hfill
        \subfloat[3DGS~\cite{kerbl20233d} (RGB)]{\includegraphics[width=\wp]{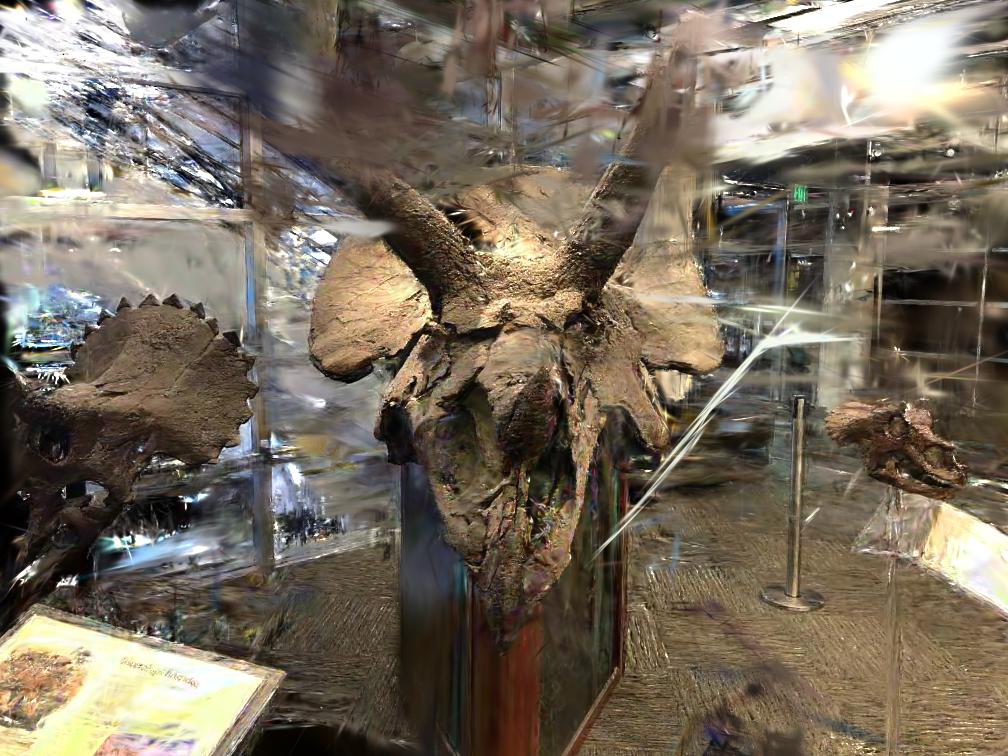}}
        \hfill
        \subfloat[3DGS~\cite{kerbl20233d} (Depth)]{\includegraphics[width=\wp]{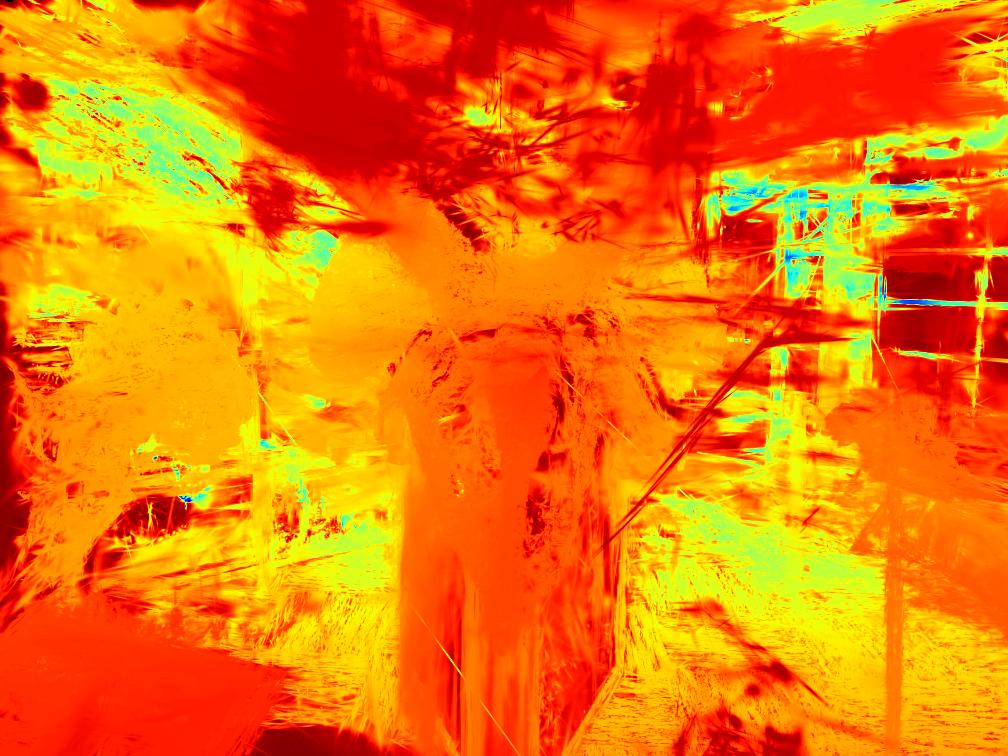}}
        \hfill
        \subfloat[Ours (RGB)]{\includegraphics[width=\wp]{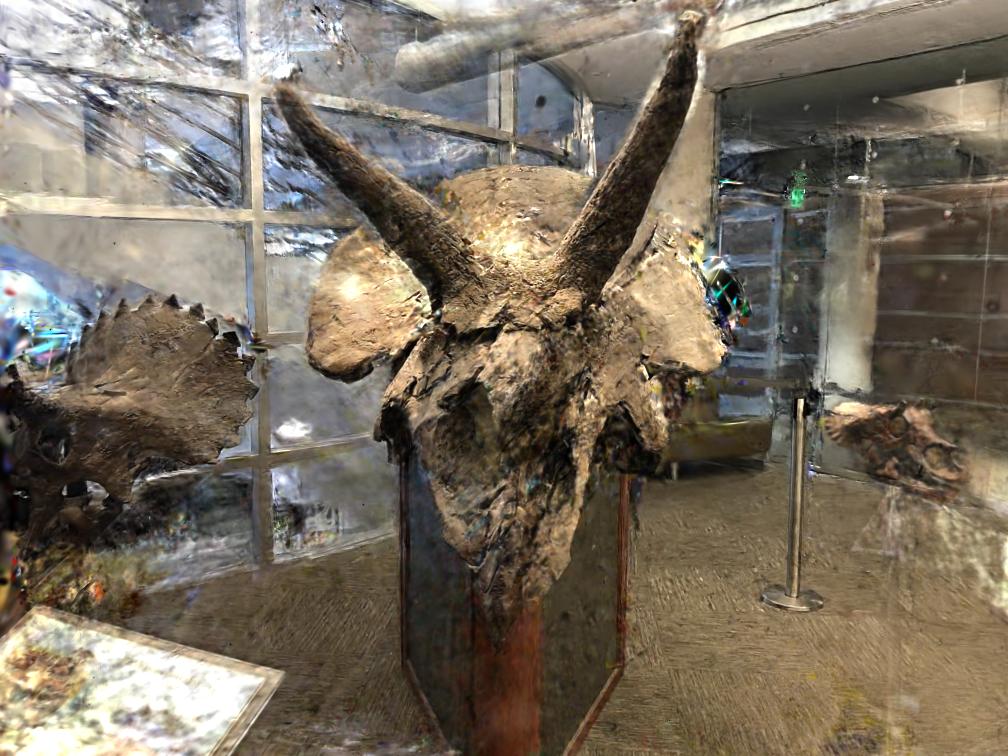}}
        \hfill
        \subfloat[Ours (Depth)]{\includegraphics[width=\wp]{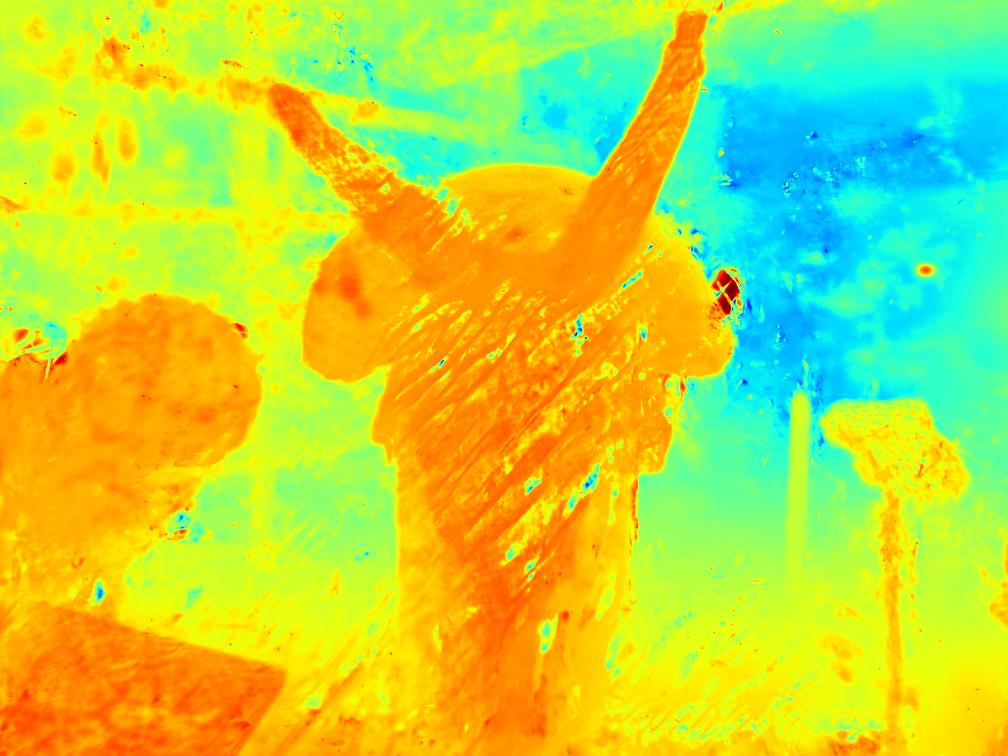}}
        \\
    \end{minipage}
    \caption{\textbf{Qualitative comparison in NeRF-LLFF~\cite{mildenhall2019llff} dataset.}
        We visualize the distinction between 3D Gaussian Splatting (3DGS)~\cite{kerbl20233d} and our method in both 2-view and 5-view settings. 
        Driven primarily by color loss, 3DGS struggled to achieve desirable geometry.
        Our approach consistently established plausible geometric structures with depth guidance, resulting in superior reconstruction outcomes.
    }
    \vspace{-1mm}
    \label{fig:comparison_llff}
\end{figure*}
\begin{figure}[t!]
    \captionsetup{font=small} 
        \newcommand{\hp}{7.65cm}
	\centering
	\subfloat[]{\includegraphics[height=\hp]{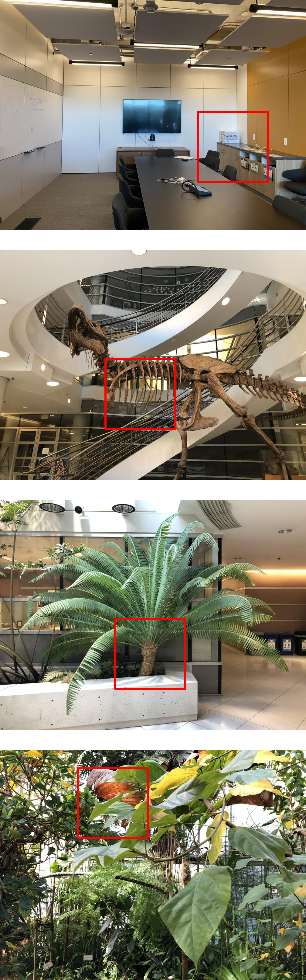}}
        \vspace{-0.05in}
        \hfill
	\centering
	\subfloat[]{\includegraphics[height=\hp]{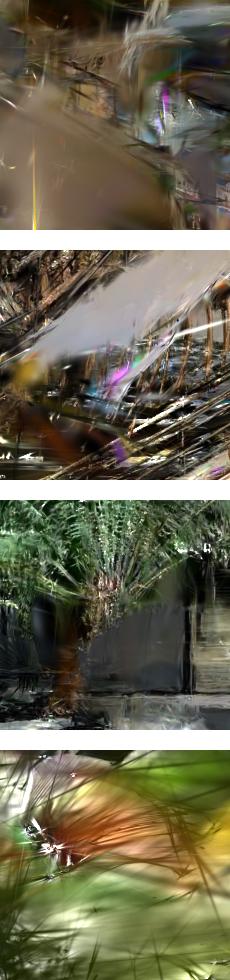}}
	\vspace{-0.05in}
        \hfill
	\centering
        \subfloat[]{\includegraphics[height=\hp]{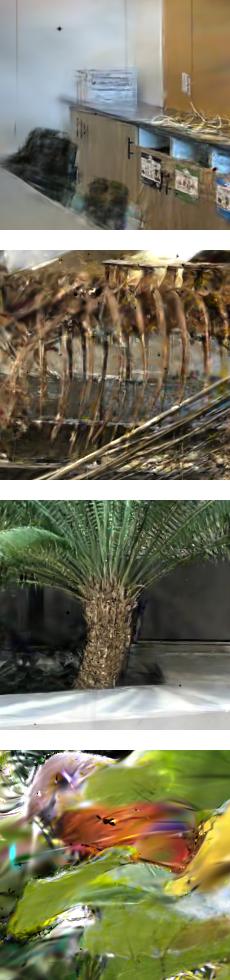}}
	\vspace{-0.05in}
        \hfill
	\centering
        \subfloat[]{\includegraphics[height=\hp]{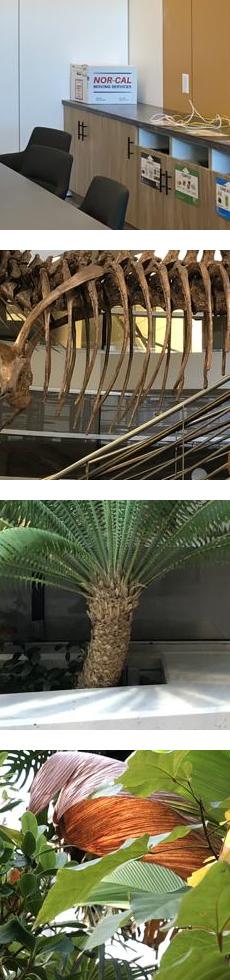}}
	\vspace{-0.05in}
        \hfill
	\vspace{-0.05in}
\vspace{3mm}
\caption{\textbf{Details in cropped patches.} 
(a) Input View (b) 3DGS~\cite{kerbl20233d} (c) Ours (d) Ground Truth.
Our method produces superior reconstruction results compared to 3DGS~\cite{kerbl20233d}, leveraging additional geometric cues. 
Our method establishes stable geometry, outperforming 3DGS in reconstruction quality.}
\label{fig:cropbox}
\vspace{-1.4 em}
\end{figure}

\paragraph{Implementation details.}
For a fair comparison among different options, it is essential to use unified coordinates in each scene and standardize the evaluation values. 
We achieve this by processing the entire images of a scene through COLMAP to obtain consistent camera poses and feature points, selecting those relevant to each k-shot experiment.
We select $k$ cameras from the train set and extract the feature points that are visible in at least three out of the $k$ cameras.
We use these feature points as depth guidance $D_{\text{sparse}}$ in \eqnref{fitting_estdepth} and initial points for Gaussian splatting optimization.
In the baseline(3DGS), we use the same $k$ camera poses and the same filtered initial points, reporting the evaluation values at 30k iterations like in the original setup. 
For the \textit{oracle}, we aim to illustrate the effectiveness of precise depth. 
We create pseudo-GT depth by optimizing the entire images of both the train and test.
We replace the estimated depth from our method with pseudo-GT depth and report the results as an oracle.
Lastly, we implement the differentiable depth rasterizer outlined in \eqnref{depth_render} based on CUDA.

\begin{table}[t]
    \centering
    \scalebox{0.73}{
    \begin{tabular}{c|ccc|ccc}
        \toprule[1.0pt]
        \multirow{2}{*}{Method} & \multicolumn{3}{c|}{2-views} & \multicolumn{3}{c}{5-views}  \\
         & PSNR$\uparrow$ & SSIM$\uparrow$ & LPIPS$\downarrow$ & PSNR$\uparrow$ & SSIM$\uparrow$ & LPIPS$\downarrow$ \\
        \midrule
        \midrule
        w/o Adjustment & \multirow{2}{*}{7.86} & \multirow{2}{*}{0.319} & \multirow{2}{*}{0.740} & \multirow{2}{*}{10.01} & \multirow{2}{*}{0.346} & \multirow{2}{*}{0.761} \\
        {(\secref{3_EstDepthOpt})} & & & & & & \\
        \midrule
        w/o $\mathcal{L}_{depth}$ & \multirow{2}{*}{11.49} & \multirow{2}{*}{0.344} & \multirow{2}{*}{0.533} & \multirow{2}{*}{12.97} & \multirow{2}{*}{0.506} & \multirow{2}{*}{0.418} \\
        (\secref{3_RenderDepthFit}) & & & & & & \\
        \midrule
        w/o $\mathcal{L}_{smooth}$ & \multirow{2}{*}{14.75} & \multirow{2}{*}{0.415} & \multirow{2}{*}{0.391} & \multirow{2}{*}{17.79} & \multirow{2}{*}{0.561} & \multirow{2}{*}{0.297} \\
        (\secref{3_DepthSmooth}) & & & & & & \\
        \midrule
        w/o early stop & \multirow{2}{*}{13.99} & \multirow{2}{*}{0.345} & \multirow{2}{*}{0.433} & \multirow{2}{*}{17.28} & \multirow{2}{*}{0.494} & \multirow{2}{*}{0.333} \\
        (\secref{3_OptSet}) & & & & & & \\
        \midrule 
        \midrule
        Ours & 15.91 & 0.420 & 0.362 & 18.39 & 0.565 & 0.296 \\
        \bottomrule[1.0pt]
    \end{tabular}
    }
    \vspace{-0.6 em}
    \caption{\textbf{Ablations.}
    We describe the ablation studies on each element of the proposed method. 
    We also present an experiment supervising with sparse depth from COLMAP instead of dense depth.
    The reported values are evaluated in \textit{Horns}.
    }
    \label{tab:ablation}
    \vspace{-0.4 em}
\end{table}
\begin{table}[t]
    \centering
    \scalebox{0.68}{
    \begin{tabular}{c|ccc|ccc}
        \toprule[1.0pt]
        Initialization & \multicolumn{3}{c|}{2-views} & \multicolumn{3}{c}{5-views}  \\
        points & PSNR$\uparrow$ & SSIM$\uparrow$ & LPIPS$\downarrow$ & PSNR$\uparrow$ & SSIM$\uparrow$ & LPIPS$\downarrow$ \\
        \midrule
        \midrule
        COLMAP from & \multirow{2}{*}{19.80} & \multirow{2}{*}{0.567} & \multirow{2}{*}{0.232} & \multirow{2}{*}{23.72} & \multirow{2}{*}{0.740} & \multirow{2}{*}{0.144} \\
        sparse-view (Ours) & & & & & & \\
        \midrule
        Unprojected & \multirow{2}{*}{16.39} & \multirow{2}{*}{0.457} & \multirow{2}{*}{0.281} & \multirow{2}{*}{19.15} & \multirow{2}{*}{0.569} & \multirow{2}{*}{0.222} \\
        from $D^{\ast}_{\text{dense}}$ & & & & & & \\
        \midrule
        COLMAP & \multirow{2}{*}{21.18} & \multirow{2}{*}{0.681} & \multirow{2}{*}{0.200} & \multirow{2}{*}{24.09} & \multirow{2}{*}{0.778} & \multirow{2}{*}{0.127} \\
        from all-view & & & & & & \\
        \bottomrule[1.0pt]
    \end{tabular}
    }
    \vspace{-0.6 em}
    \caption{\textbf{Comparison of Initialization Methods.}
    We describe the ablation studies on each element of the proposed method. 
    We also present an experiment supervising with sparse depth from COLMAP instead of dense depth.
    The reported values are evaluated in \textit{Fortress}.
    }
    \label{tab:initialization}
    \vspace{-1.4 em}
\end{table}

\subsection{Experiment results}
\begin{figure*}[t!]
    \centering
    \includegraphics[width=\textwidth, height=0.3\textwidth]{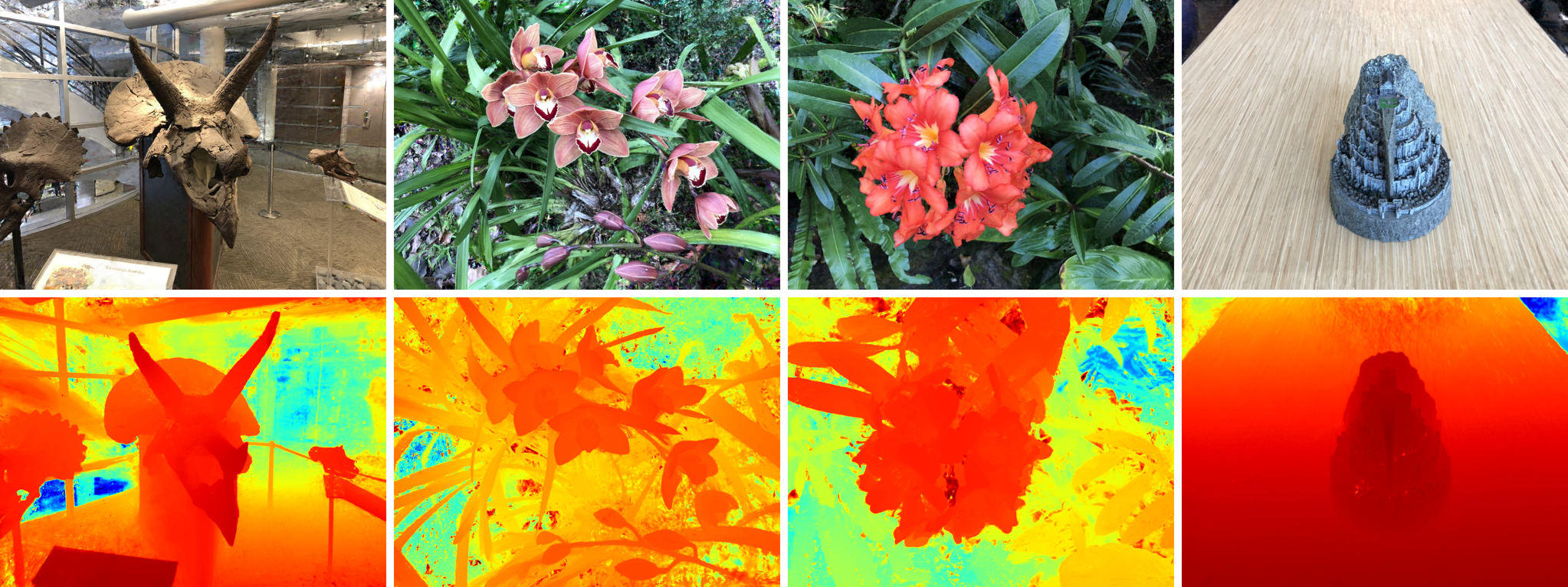} 
    \vspace{-0.6cm}
    \caption{\textbf{Example results utilizing pseudo-GT depth (\textit{oracle}).} 
    Accurate depth facilitates high-quality 3D reconstruction, even with a limited number of images.
    Fine details are perceptible in both RGB and depth.
    }
    \label{fig:oracle}
    \vspace{-0.4cm}
\end{figure*}

We present the comparison results of 3DGS, our method, and oracle for NeRF-LLFF scenes in \tabref{result_llff}.
Across all methods and scenes, a decrease in the number of used images consistently results in lower visual quality.
Our method typically demonstrates superior results compared to 3DGS, particularly when the number of images is limited.
\figref{comparison_llff} visualize the difference between 3DGS and our method.
The depth map highlights the geometry failure of 3DGS in the few-shot case. 
For instance, in the 2-view of the \textit{Fern}, it displays entirely erroneous geometry compared to the similarity in RGB.
In har conditions like the 2-view scenario, 3DGS often fails to form appropriate geometry.
In contrast, our method forms plausible geometry while generating an attractive image.
We present additional examples in \figref{cropbox}. The cropped patches demonstrate that our method achieves better results through depth guidance.
Hence, we confirm that the geometric cues provided by depth become significantly beneficial for the reconstruction in Gaussian splatting, especially when the number of images is limited.
This fact is reaffirmed by the remarkably high performance of the oracle, which employs accurate geometry.
The example images of the oracle demonstrate the effectiveness of accurate depth, as depicted in \figref{oracle}.
The rich information provided by pseudo-GT depth enables the creation of detailed and reliable results even with a limited number of images.

An important observation to note is the substantial reliance of our approach on the pre-trained monocular depth estimation model.
We exploit the pre-trained model of ZoeDepth~\cite{bhat2023zoedepth} trained on the indoor dataset NYU Depth v2~\cite{silberman2012indoor} and urban dataset KITTI~\cite{menze2015object}.
As a result, our model reports relatively higher performance in indoor scenes (Fortress, Room, Fern) and comparatively worse results for natural scenes (Orchids, Flower).
Note that the \textit{Leaves} presents challenges for COLMAP, leading to generally unsuccessful Gaussian splatting training.

\subsection{Ablations}
We present ablation studies for each component of our proposed method in \tabref{ablation}.
The first and second rows demonstrate the necessity of absolute depth guidance.
Without the adjustment process in \secref{3_EstDepthOpt}, the dense Depth $D_{\text{dense}}$ has an incorrect scale from the monocular depth estimation model. 
The depth is misaligned with the camera intrinsic parameters from COLMAP, leading to complete training failure.
We also observed optimization failure when solely utilizing unsupervised smooth constraints without depth supervision introduced in \secref{3_RenderDepthFit}.
The application of smoothness constraints without absolute geometry supervision yields worse results compared to the baseline.
The third and fourth rows of \tabref{ablation} demonstrate the degree of performance enhancement from additional techniques.
With the depth supervision $D^{\ast}_{\text{dense}}$, the smoothness constraints in \secref{3_DepthSmooth} contribute to performance improvement by providing additional geometric cues.
Notably, the early stop mechanism introduced in \secref{3_OptSet} plays a pivotal role in preventing performance degradation within our approach. 
By leveraging depth loss, it scrutinizes the divergence of splats from the prescribed geometry guide, effectively halting potential instances of overfitting.

In \tabref{initialization}, we compared the performances between the different Gaussian splatting initializations.
The second row illustrates the outcomes when utilizing a point cloud produced by unprojecting dense depth $D^{\ast}_{\text{dense}}$ as initialization points. 
The numerous initial points generated through unprojection are not effectively merged or pruned, resulting in lower performance compared to the sparse COLMAP initialization.
In contrast, the outcomes depicted in the third row assume the utilization of all COLMAP points.
Employing a multitude of favorable initial points unattainable with k images contributes to the enhancement of outcomes via dense depth adjustment and initialization.

\section{Limitation and Future Work}
Our approach demonstrated the feasibility of Gaussian splatting optimization in a few-shot setting through depth guidance, yet it has limitations. 
Firstly, it heavily relies on the estimation performance of the monocular depth estimation model.
Moreover, this model's depth estimation performance can vary based on the learned data domain, consequently affecting the performance of Gaussian splatting optimization. 
Additionally, relying on fitting the estimated depth to COLMAP points means a dependency on COLMAP's performance, rendering it incapable of handling textureless plains or challenging surfaces where COLMAP might fail.
We leave as future work the optimization of 3D scenes by interdependent estimated depths rather than COLMAP points. 
Also, exploring methods to regularize geometry across various datasets, including areas where depth estimation, such as the sky, might be challenging, is another avenue for future work.

\section{Conclusion}
In this work, we introduce Depth-Regularized Optimization for 3D Gaussian Splatting in Few-Shot Images, a model for learning 3D Gaussian splatting with a small number of images.
Our model regularizes the splats using depth, demonstrating the effectiveness of such geometric guidance. 
To acquire dense depth guidance, we exploit a monocular depth estimation model and adjust the depth scale based on SfM points.
We examined the effectiveness of our proposed depth loss, unsupervised smooth constraint, and early stop technique in the NeRF-LLFF dataset.
Our method outperforms 3D Gaussian splatting in a few-shot setting, creating plausible geometry.
Finally, we demonstrated through additional experiments that improved depth and initialization points significantly enhance the performance of Gaussian splatting-based 3D reconstruction.

\clearpage
{
    \small
    \bibliographystyle{ieeenat_fullname}
    \bibliography{main}
}


\end{document}